\documentclass[letter,twocolumn,twoside]{article}

\usepackage{xspace}
\usepackage{times}
\usepackage{amsfonts}
\usepackage{verbatim}
\usepackage{graphicx}     
\usepackage{subfigure}   
\usepackage{multirow}     
\usepackage{psfrag,psfig,epsfig,epsf}
\usepackage{graphics}
\usepackage{latexsym,amsfonts,amssymb}
\usepackage[authoryear,round]{natbib}
\usepackage[noline,noend,ruled,linesnumbered]{algorithm2e}
\usepackage{amsmath}
\usepackage{fancyhdr}
\usepackage{fullpage}
\usepackage{wrapfig}
\usepackage{subfigure}

\renewcommand*{\cite}{\citep}

\begin{document}

\title{Sampling-based Roadmap Planners are Probably Near-Optimal after
  Finite Computation}
\author{Andrew Dobson \and George V. Moustakides \and
Kostas E. Bekris}
%
%
%


\newcommand{\prm}{\ensuremath{\tt PRM}}
\newcommand{\sprm}{\text{s-}\ensuremath{\tt PRM}}
\newcommand{\prmstar}{\ensuremath{{\tt PRM}^*}}
\newcommand{\kprmstar}{\ensuremath{{\tt k}\text{-}{\tt PRM}^*}}
\newcommand{\srs}{\ensuremath{{\tt SRS}}}
\newcommand{\irs}{\ensuremath{{\tt IRS}}}
\newcommand{\rrg}{\ensuremath{{\tt RRG}}}
\newcommand{\rrt}{\ensuremath{{\tt RRT}}}
\newcommand{\rrtstar}{\ensuremath{{\tt RRT}^*}}
\newcommand{\spars}{\ensuremath{{\tt SPARS}}}
\newcommand{\pno}{\ensuremath{{\tt PNO}}}
\newcommand{\pac}{\ensuremath{{\tt PAC}}}
\newcommand{\pnoprm}{\ensuremath{{\tt PNO}\text{-}{\tt PRM^*}}}

\newcommand{\cspace}{\ensuremath{\mathcal{C}}}
\newcommand{\cfree}{\ensuremath{\mathcal{C}_\text{free}}}
\newcommand{\cinv}{\ensuremath{\mathcal{C}_\text{obs}}}

\newcommand{\reals}{\ensuremath{\mathbb{R}}}
\newcommand{\expect}{\ensuremath{\mathbb{E}}}
\newcommand{\pr}{\ensuremath{\mathbb{P}}}
\newcommand{\cover}{\ensuremath{\Omega_n^\beta}}
\newcommand{\ccover}{\ensuremath{\Omega^C}}
\newcommand{\pcover}{\pr(\cover)}
\newcommand{\zmu}{\ensuremath{y}}
\newcommand{\success}{\ensuremath{\pr_{success}}}
\newcommand{\pdes}{\ensuremath{\pr_{DES}}}
\newcommand{\ddes}{\ensuremath{\delta_{DES}}}

\newcommand{\ballrad}{\ensuremath{\beta_n}}
\newcommand{\balls}{\ensuremath{M_n}}
\newcommand{\ballnaught}{\ensuremath{M_0}}
\newcommand{\clear}{\ensuremath{\epsilon_n}}
\newcommand{\enaught}{\ensuremath{{\epsilon_0}}}
\newcommand{\connect}{\ensuremath{r_n}}
\newcommand{\lan}{\ensuremath{\lambda_n}}
\newcommand{\boundn}{\ensuremath{\delta_n}}
\newcommand{\nnaught}{\ensuremath{n_0}}
\newcommand{\gpno}{\ensuremath{\gamma_\pno}}
\newcommand{\gprm}{\ensuremath{\gamma_{\prmstar}}}
\newcommand{\ugly}{\ensuremath{\chi}}
\newcommand{\ougly}{\ensuremath{\psi}}
\newcommand{\lopt}{\ensuremath{I_{\clear}^*}}
\newcommand{\lnaupt}{\ensuremath{I^*_{\epsilon_0}}}
\newcommand{\lstar}{\ensuremath{I^*}}
\newcommand{\lalg}{\ensuremath{I_n}}
\newcommand{\lnaught}{\ensuremath{I_{0}}}
\newcommand{\lspan}{\ensuremath{I_{span}}}
\newcommand{\lm}{\ensuremath{I_{m}}}
\newcommand{\lk}{\ensuremath{I_{m+1}}}
\newcommand{\lkk}{\ensuremath{I_{k}}}
\newcommand{\lj}{\ensuremath{I_{m+2}}}
\newcommand{\lo}{\ensuremath{I_{1}}}
\newcommand{\lt}{\ensuremath{I_{2}}}
\newcommand{\lr}{\ensuremath{I_{3}}}
\newcommand{\volcon}{\ensuremath{V}}
\newcommand{\voldiff}{\ensuremath{dVol}}
\newcommand{\sint}{\ensuremath{S}}
\newcommand{\cd}{\ensuremath{{\frac{(d-1)}{(d+2)}}}}

\newcommand{\ehalf}{\ensuremath{{\epsilon \over 2}}}
\newcommand{\tehalf}{\ensuremath{{3 \epsilon \over 2}}}
\newcommand{\efourth}{\ensuremath{{\epsilon \over 4}}}
\newcommand{\segs}{\ensuremath{{\lceil \frac{\lopt}{\clear} \rceil}}}
\newcommand{\newm}{\ensuremath{{\big \lceil {\lopt \over \epsilon} \big \rceil}}}
\newcommand{\ratio}{\ensuremath{\frac{|\ball(r)|}{|\cfree|}}}
\newcommand{\eratio}{\ensuremath{\frac{|\ball(\epsilon)|}{|\cfree|}}}

\newcommand{\piopt}{\ensuremath{\pi^*_{\clear}}}
\newcommand{\pinaupt}{\ensuremath{\pi^*_{0}}}
\newcommand{\pinaught}{\ensuremath{\pi_{0}}}
\newcommand{\ball}{\ensuremath{\mathcal{B}}}
\newcommand{\local}{\ensuremath{\mathit{L}}}

\newtheorem{theorem}{Thm.}
\newtheorem{lemma}{Lemma}
\newtheorem{corollary}{Corr.}
\newtheorem{property}{Prop.}
\newtheorem{definition}{Defn.}
\newtheorem{assumption}{Assumption}
\newcommand{\qed}{\nobreak \ifvmode \relax \else
      \ifdim\lastskip<1.5em \hskip-\lastskip
      \hskip1.5em plus0em minus0.5em \fi \nobreak
      \vrule height0.75em width0.5em depth0.25em\fi}

\newenvironment{myitem}{\begin{list}{$\bullet$}
{\setlength{\itemsep}{-0pt}
\setlength{\topsep}{0pt}
\setlength{\labelwidth}{0pt}
\setlength{\leftmargin}{10pt}
\setlength{\parsep}{-0pt}
\setlength{\itemsep}{0pt}
\setlength{\partopsep}{0pt}}}%
{\end{list}}

\maketitle

\begin{abstract}
Sampling-based motion planners have proven to be efficient solutions
to a variety of high-dimensional, geometrically complex motion
planning problems with applications in several domains.  The
traditional view of these approaches is that they solve challenges
efficiently by giving up formal guarantees and instead attain
asymptotic properties in terms of completeness and optimality.  Recent
work has argued based on Monte Carlo experiments that these approaches
also exhibit desirable probabilistic properties in terms of
completeness and optimality after finite computation. The current
paper formalizes these guarantees. It proves a formal bound on the
probability that solutions returned by asymptotically optimal
roadmap-based methods (e.g., \prmstar) are within a bound of the
optimal path length $\lopt$ with clearance $\epsilon_n$ after a finite
iteration $n$. This bound has the form $\pr\big( |\lalg - \lopt|
\leq \delta\cdot\lopt \big) \leq \success$, where $\delta$ is an error
term for the length a path in the \prmstar\ graph, $\lalg$. This bound
is proven for general dimension Euclidean spaces and evaluated in
simulation. A discussion on how this bound can be used in practice, as
well as bounds for sparse roadmaps are also provided.

\end{abstract}

\section{Background}

Early contributions in sampling-based motion planning focused on
overcoming the computational challenges posed by motion planning
problems with high dimensionality and geometrically complex spaces
\cite{ Latombe1991Robot-Motion-Pl, LaValle2006Planning-Algori,
  Choset2005Principles-of-R}.  Two alternative families of sampling
based planners emerged during this process, roadmap-based methods,
such as \prm \cite{Kavraki1996Probabilistic-R,Kavraki1998prm}, which
are suited to multi-query planning, and tree-based approaches, such as
\rrt \cite{LaValle1998RRT, LaValle2000Rapidly-explore}.  Formal
analysis of these methods followed, showing they are probabilistically
complete \cite{Kavraki1998Analsis-of-Prob,Hsu1998On-finding-narr,
  Ladd2004Measure-Theoret, Chaudhuri2009Smoothed}.  Though these
methods are probabilistically complete, the literature has shown that
solution non-existence can be detected under certain conditions
\cite{Varadhan2007Star-shaped-Roa, McCarthy2012Proving-path-no}. Other
work tries to return high clearance paths, or characterize the
$\cspace$-space obstacles
\cite{Wilmarth1999MAPRM:-A-probab,Amato1998OBPRM:-An-Obsta}, and
others return high quality solutions in practice
\cite{Raveh2011PathMerging}.

A major recent breakthrough was the identification of the conditions
under which these methods asymptotically converge to optimal paths
\cite{Karaman2011Sampling-based-, Karaman2010Incremental-Sam},
resulting in algorithms such as \rrtstar and \prmstar.  Both
probabilistic completeness and asymptotic optimality relate to
desirable properties after infinite computation time.  Since these
methods are practically terminated after some finite amount of
computation, these guarantees cannot provide information about
expected path cost or of solution non-existence in practice \cite{
  Varadhan2007Star-shaped-Roa,
  McCarthy2012Proving-path-no}. Nevertheless, experiments show that
asymptotically optimal methods do have very good behavior in terms of
path quality after finite computation time, even when optimality
constraints are relaxed to create more efficient methods with path
length guarantees \cite{Marble2013ANOJournal,
  Salzman2013Asymptotically-, Wang2013A-fast-streamin}. To address the
gap between practical experience and formal guarantees, recent work by
the authors has proposed that asymptotically optimal sampling-based
planners also exhibit \emph{probabilistic near-optimality} properties
after finite computation using Monte Carlo experiments
\cite{Dobson2013A-Study-on-the-}. This kind of guarantee is similar to
the concept of Probably Approximately Correct (\pac) solutions in the
machine learning literature \cite{Valiant84}. The focus in this work
is on the properties of roadmap-based methods, such as \prmstar, as
they are easier to analyze.

This work formally shows the \emph{Probabilistic Near-Optimality}
(\pno) of sampling-based roadmap methods in general settings and 
with limited assumptions.  It provides the
following contributions relative to the state-of-the-art and the
previous contribution by the authors \cite{Dobson2013A-Study-on-the-}:
\begin{myitem}
\item Prior work relied on Monte Carlo simulations to provide path
  length bounds, while this work achieves tight, closed-form bounds.
  This required solving a problem in geometric probability, which to
  the best of the authors' knowledge had not been addressed before.
\item The framework is extended to work with a version of
  \prmstar\ which constructs a roadmap having $O(n\log{n})$ edges,
  which is in the order of the lower bound for asymptotic
  optimality. Prior work used a method called \pnoprm, which creates
  $O(n^2)$ edges.
\end{myitem}

\section{Problem Setup}\label{sec:setup}

\noindent This section introduces terminology and definitions 
required for the formal analysis.  This work examines kinematic planning
in the configuration space, $\cspace$, where a robot's configuration
$q \in \cfree$ is cast as a point.  $\cspace$ is partitioned 
into the collision free (\cfree) and colliding (\cinv)
configurations.  This work reasons over \cspace\ as a metric space, 
using the Euclidean $L_2$-norm as a distance metric.
The objective is to compute a path $\pi_n: [0,1] \to \cfree$ after
finite iterations $n$ with path length guarantees relative to an 
$\epsilon$-robust feasible path, i.e.
a path with minimum distance to \cinv\ of at least $\epsilon$. If a 
motion planning problem is robustly feasible, there exists a 
set of $\epsilon$-robust paths which answer a query, $(q_{start},
q_{goal})$.  Let the path of minimum length from the set
be denoted as $\piopt$, with length $\lopt$.  The path planning problem 
this work considers is the following:

\begin{definition}[Robustly Feasible Motion Planning] 

\noindent Let the tuple $(\cspace, q_{start}, q_{goal}, \enaught)$
be an instance of a Robustly Feasible Motion Planning Problem.
Given a configuration space $\cspace$, two configurations
$q_{start}$, $q_{goal} \in \cfree \subset \cspace$ and a clearance
value $\enaught$ so that an $\enaught$-robust path
$\pi_{\enaught}$ exists so that $\pi_{\enaught}(0) = q_{start}$
and $\pi_{\enaught}(0) = q_{goal}$, find a solution path $\pi$ so
that $\pi(0) = q_{start}$ and $\pi(1) = q_{goal}$.
\end{definition}

To solve this problem, a slight variation of the \prmstar\ algorithm
is applied \cite{Karaman2011Sampling-based-}.
The high-level operations of \prmstar\ are as follows:
\begin{itemize}
\item \prmstar\ generates configurations $Q$ in \cfree, rejecting samples
generated in \cinv, and then adding $Q$ to a graph, $G=(V,E)$, i.e. 
$V \gets V \cup Q$. 
\item For each sample, an $\connect$ radius, where $\connect = 
\gamma_{\prmstar} \cdot \Big( \frac{\ln{n}}{n} \Big)^{\frac{1}{d}}$ 
\cite{Karaman2011Sampling-based-}, local neighborhood 
in \cfree\ is examined.  If a local path to a neighbor can 
be generated which remains entirely in \cfree, an edge connecting them 
is added to $E$.
\item The above steps are repeated iteratively until some stopping
criterion is met.
\end{itemize}

\noindent This work's variant \prmstar\ uses a larger
connection radius, $\gamma_{\pno} = 2\cdot\gamma_{\prmstar}$,
and the reason why becomes apparent from the analysis.  The larger
connection radius allows for the following property to be argued:

\begin{property}[Probabilistic Near-Optimality for RFMP] 
\noindent An algorithm $ALG$ is probabilistically near-optimal for an
RFMP problem $(\cspace, q_{start}, q_{goal}, \enaught)$, if for a
finite iteration $n > \nnaught$ of $ALG$ and a given error threshold
$\delta$, it is possible to compute a probability $\success$ so that
for the length $I_n$ of a path $\pi_n$ answering the query in the 
planning structure computed by $ALG$ at iteration $n$:
$$\pr(|\lalg - \lopt| > \delta\cdot \lopt)< 1-\success $$ where
$\lopt$ is the length of the optimum $\epsilon_n$-robust path $\piopt$
for a value $\clear < \enaught$.
\end{property}

\begin{figure}[ht]
\includegraphics[width=2.95in]{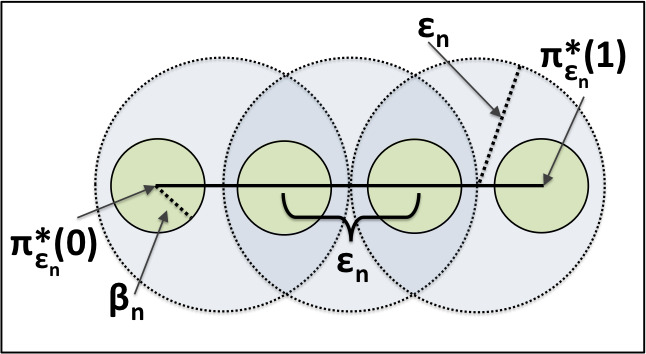}
\centering
\vspace{-0.1in}
\caption{Hyperballs over an optimal path with radius $\ballrad$ and
separation $\clear$.  Consecutive balls lie entirely within
some clearance ball $\ball_{\clear}(\piopt(\tau_t))$.}
\label{fig:ballstruct}
\end{figure}
\noindent The clearance of the optimum path $\clear$ considered at 
iteration $n$ and the iteration $n_0$ after which point the guarantee
can be achieved, can be computed given the analysis in this work.

Probabilistic Near-Optimality (\pno) can be argued by reasoning over 
a theoretical construction of hyperballs tiled over \piopt, where 
hyperballs are denoted as $\ball_r(q_c)$, being centered at 
configuration $q_c$ and having radius $r$.  The construction of these 
hyperballs is illustrated in Figure \ref{fig:ballstruct}.  
Construct $\balls + 1 = \segs +
1$ balls, $B_n$ centered along $\piopt$, i.e. $B_n = \{ \ball_{\ballrad}
(\piopt(\tau_0)), \ldots, \ball_{\ballrad}(\piopt(\tau_{\balls}))\}$,
having radius $\ballrad \leq \frac{1}{2}\clear$, where $\clear
= \frac{1}{2}\connect$, and where $\connect$ is the connection radius 
used by the algorithm.  The construction enforces the centers of the balls
to be $\clear$ apart, and by choice of $\ballrad$, these balls have
empty intersections.  Then, since $\connect \geq 4\ballrad$, the
algorithm will attempt connections between any pairs of points between
consecutive hyperballs.  \pno\ guarantees are over a path in the 
planning structure with length \lalg.  This path corresponds to the
set of all the first samples generated in each of the hyperballs.

Then, using the steps from related work 
\cite{Karaman2011Sampling-based-}, $\gpno$ can be 
derived, as well as, $k(n)$ for an equivalent \kprmstar\ 
variant.  These values are derived in the next section.

\section{Derivation}\label{sec:derive}

This section provides a bound on the probability 
that \prmstar\ returns poor-quality paths.  Namely, it constructs
the probability of $\lalg$ being $1 + \delta$ times larger than the 
optimal path after $n$ iterations.  Then, it provides a guarantee
$ \pr ( |\lalg - \lopt| > \delta\cdot \lopt)< 1-\success $, where 
$\delta > 0$ is an input multiplicative 
bound, and $\success \in (0,1)$ is a confidence bound.  A guarantee 
of this type can be considered a Probably Near Optimal (\pno) 
property.  First, the algorithmic parameters $\gpno$ and $k(n)$ are 
derived.

\subsection{Deriving $\gpno$}
\label{sec:gamma}

This section employs the same steps as the derivation for $\gprm$ in
the literature \cite{Karaman2011Sampling-based-}.
The objective of this section is to leverage a bound on the probability 
that \prmstar\ will fail to produce a sample in each of the hyperballs 
over \piopt\ to derive an appropriate constant for the 
connection radius. Let this connection radius employed by the \prmstar\ 
variant be $\connect = \gpno \Big( \frac{\ln{n}}{n} 
\Big)^{\frac{1}{d}}$. Then, by construction, this connection radius 
is at least four times larger than the radius of a hyperball, 
i.e. $\ballrad < \frac{1}{4}\gpno \Big( \frac{\ln{n}}{n} 
\Big)^{\frac{1}{d}}$.  Then,

$$|\ball_{\ballrad}| = \volcon_d\ballrad^d < \volcon_d\big(\frac
{\connect}{4} \big)^d = \volcon_d\cdot\frac{\ln{n}}{n}\big( \frac
{\gpno}{4}\big)^d , $$

\noindent where $\volcon_d = |\ball_1|$ is the $d$-dimensional 
constant for the volume of a hyperball.  Also by construction,
$\clear \geq \frac{1}{2}\gpno \Big( \frac{\ln{n}}{n} 
\Big)^{\frac{1}{d}}$.  Then, the number of hyperballs constructed
over \piopt\ can be bounded by $\balls \leq \frac{\lopt}{\clear}
= \frac{2\lopt}{\gpno}\Big( \frac{n}{\ln{n}} \Big)^{\frac{1}{d}}$.

Then, in line with previous work in the literature 
\cite{Kavraki1998Analsis-of-Prob,Karaman2011Sampling-based-}, the 
probability of failure can be bounded using the probability that a 
single hyperball contains no sample.  The event that a single 
hyperball does not contain a sample is denoted as $\mathcal{F}$, and 
has probability:

$$\pr(\mathcal{F}) = \bigg( 1 - \frac{|\ball_{\ballrad}|}{|\cfree|}\bigg)^n
< \bigg(1- \frac{\volcon_d}{|\cfree|}\cdot\frac{\ln{n}}{n}\cdot \Big(
\frac{\gpno}{4}\Big)^d \bigg)^n $$

\noindent Then, since $(1 - x)^t \leq e^{-tx}$,

\begin{equation}
\pr(\mathcal{F}) \leq e^{-\frac{\volcon_d}{|\cfree|}\cdot\ln{n}\cdot \big(
\frac{\gpno}{4}\big)^d } = n^{-\frac{\volcon_d}{|\cfree|}\cdot \big(
\frac{\gpno}{4}\big)^d} 
\label{eq:pfail}
\end{equation}

Now, compute bounds on the event \ccover\ that at least one ball does 
not contain a sample as:

$$\pr(\ccover) = \pr\big(\bigcup_{\balls} \mathcal{F} \big) \leq 
\sum_{i=1}^{\balls} \pr(\mathcal{F}) = (\balls)\pr(\mathcal{F})$$

\noindent Substituting the computed value for \balls, and 
$\pr(\mathcal{F})$ from Eq. \ref{eq:pfail}:

\begin{multline*}
\pr(\ccover) \leq \bigg(\frac{2\lopt}{\gpno}\Big( \frac{n}{\ln{n}} 
\Big)^{\frac{1}{d}}\bigg)n^{-\frac{\volcon_d}{|\cfree|}\cdot \big(
\frac{\gpno}{4}\big)^d} =\\ \bigg(\frac{2\lopt}{\gpno} \frac{1}
{(\ln{n})^{\frac{1}{d}}}\bigg)n^{-\big(\frac{\volcon_d}{|\cfree|}
\cdot \big(\frac{\gpno}{4}\big)^d - \frac{1}{d}\big)}
\end{multline*}

\noindent Now, if $\sum_{i=1}^\infty\pr(\ccover)$ is less than infinity,
this implies by the Borel-Cantelli theorem that $\pr(
\limsup_{n\to\infty} \ccover) = 0$ \cite{Grimmet2001Probability-and}.
Furthermore, by the Zero-one Law, $\pr( \limsup_{n\to\infty} \ccover) 
= 0 \Rightarrow \pr(\limsup_{n\to\infty} \cover) = 1$, meaning the
probability of coverage converges to $1$ in the limit.

In order for the sum to be less than infinity, it is sufficient to 
show that the exponent, $\frac{\volcon_d}{|\cfree|} \cdot 
\big(\frac{\gpno}{4} \big)^d - \frac{1}{d} < 1$.  The algorithm can 
ensure this by using an appropriate value of $\gpno$.  Solving the 
inequality for $\gpno$ shows that it suffices that:
$$ \gpno > 4\bigg( \Big( 1 + \frac{1}{d}\Big)\Big( \frac{|\cfree|}
{\volcon_d} \Big) \bigg)^{\frac{1}{d}} $$

\subsection{Deriving $k(n)$ for \kprmstar}
\label{sec:kprm}

This section employs the same steps as the derivation for $k(n)$ in
the literature \cite{Karaman2011Sampling-based-}.
The objective of this section is to derive the function, $k(n)$, for
a \pno\ variant of $\kprmstar$.  The high-level idea is that it will
be shown that two events happen infinitely often with the given $k(n)$;
the set of hyperballs each contain at least one sample, and that
each ball of radius $\clear$ has no more than $k(n)$ samples inside it.
From this, it is clear that if \kprmstar\ attempts to connect each 
sample with $k(n)$ neighbors, it will attempt connections between 
samples in neighboring hyperballs.

Then, using the computed value of \gpno\ from above,
\begin{multline*}
\balls \leq \frac{1}{2}\lopt \bigg( \Big( \frac{\volcon_d}{|\cfree|} 
\Big)\Big( \frac{n}{\ln{n}} \Big)\Big( \frac{1}{1+\frac{1}{d}} \Big)
\bigg)^{\frac{1}{d}}, \text{ and } \\
|\ball(\clear)| \leq \Big(\volcon_d \cdot 4^d \Big) \Big( \frac{1}{2} 
\Big)^d \Big( 1 + \frac{1}{d} \Big) \Big( \frac{|\cfree|\ln{n}}
{n\volcon_d} \Big) =\\ 2^d\Big(1+\frac{1}{d}
\Big)\Big( \frac{|\cfree|\ln{n}}{n} \Big)
\end{multline*}

Let $A$ be an indicator random variable which takes value $1$ when
there is a sample in some arbitrary hyperball of radius $\clear$.
Then, $\expect[A] = \frac{|\ball(\clear)|}{|\cfree|} = 2^d (1 + 
\frac{1}{d})(\frac{\ln{n}}{n})$.  Since each sample is drawn 
independently of the others, the number of samples in a ball can be
expressed as a random variable $N$, such that $\expect[N] = 
n\expect[A] = 2^d(1+\frac{1}{d})\ln{n}$.  Due to $A$ being a
Bernoulli random variable, the Chernoff Bound can be employed to 
bound the probability of $N$ taking large values, namely:

$$\pr(N > (1+t)\expect[N]) \leq \bigg( \frac{e^t}{(1+t)^{(1+t)}} 
\bigg)^{\expect[N]}, t > 0$$

\noindent Then, let $t = e - 1$.  Substituting this above yields:

$$\pr(N > e\expect[N]) \leq e^{-\expect[N]} = e^{-2^d(1+
\frac{1}{d})\ln{n}} = n^{-2^d(1+\frac{1}{d})} $$

\noindent Now, in order for the $k(n)$ connections to attempt 
connections outside of a $\clear$-ball, it must be that:

$$ k(n) = k_\pno e\Big(1+\frac{1}{d}\Big) \geq 2^de\Big(1+\frac{1}{d}
\Big) = e\expect[N], $$

\noindent which clearly holds if $k_\pno = 2^d$.  This implies that
$ \pr(N > k(n)) \leq n^{-2^d(1+\frac{1}{d})} $.

Finally, consider the event $\zeta$ that even one of the balls has 
more than $k(n)$ samples:

\begin{multline*}
\pr(\zeta) = \pr\Big(\bigcup_{\balls} \pr(N > k(n)) \Big) \leq \\
\sum^\balls \pr(N > k(n)) = \balls\pr(N > k(n))
\end{multline*}

$$\pr(\zeta) \leq \frac{\lopt}{2}\bigg( \frac{\volcon_d}{\ln{n}
|\cfree|(1+\frac{1}{d})} \bigg)^{\frac{1}{d}} n^{-2^d(1+\frac{1}
{d})+\frac{1}{d}}$$

\noindent Then, it is clear that $\sum_{i=1}^\infty \pr(\zeta) < 
\infty$, which by the Borel-Cantelli Theorem implies that 
$\pr(\limsup_{n\to\infty} \zeta) = 0$, and furthermore,  $\pr(
\limsup_{n\to\infty} \zeta^C) = 1$ via the Zero-one Law, i.e. the 
number of samples in the \clear -ball is almost certainly less than 
$k(n)$.  

Finally, using the result showing the convergence of $\pr(\cover)$
to $1$, and the above result for $\pr(\zeta^C)$, it can be concluded
that $\pr(\limsup_{n\to\infty} ( \cover \cap \zeta^C ) ) = 1$, 
implying that for the choice of $k(n)$, \kprmstar\ attempts the 
appropriate connections.

\subsection{Deriving the Probability of Coverage}
\label{sec:pcover}

The derivation of the probability of path coverage leverages 
several results in the literature \cite{Kavraki1998Analsis-of-Prob,
Karaman2011Sampling-based-,Dobson2013A-Study-on-the-}.
The objective is to exactly derive the probability 
that at any finite iteration, $n$, the algorithm has generated a sample
in each of the hyperballs over \piopt.  Deriving this probability will 
work off of the result shown in prior work which gives the probability 
of coverage for a similar construction of hyperballs to that employed 
here \cite{Dobson2013A-Study-on-the-}, which shows:

\begin{equation}
\pcover = \bigg(1 - \Big(1 - \frac{|\ball_\beta|}{|\cfree|} 
\Big)^n \bigg)^{M+1}, 
\label{eq:orig_pcover}
\end{equation}

\noindent where $\beta$ is the radius of the set of hyperballs
and $M+1$ is the number of such hyperballs.  Here, the inner
term $\Big(1 - \frac{|\ball_\beta|}{|\cfree|} \Big)^n$ is the
probability of failing to throw a sample in a particular hyperball
after $n$ samples have been thrown.  Then, the probability of success
for throwing a sample in all of the hyperballs yields the above form.
This holds for any values of $\beta$ and $M$ such that
the hyperballs are disjoint, which is exactly the construction
employed in this work.  Then, substituting the values computed for
$\ballrad$ and $\balls$ from Section \ref{sec:gamma} above yields:

$$\pcover \approx \bigg(1 - \Big(1 - a \Big)^n \bigg)^{{\frac{1}{2}\lopt
\Big( b \Big)^{-\frac{1}{d}}}+1} $$

\noindent where $a = \frac{\volcon_d \Big( {\sqrt[d]{\big( 1 + 
\frac{1}{d} \big) \big( \frac{|\cfree|\ln{n}}{\volcon_dn} \big)}}
\Big)^d}{|\cfree|}$, $b = \big(1 + \frac{1}{d} \big) \big( \frac 
{|\cfree|\ln{n}} {\volcon_dn} \big)$, and $\volcon_d$ is the 
d-dimensional constant for the volume of a hyperball, i.e. 
$|\ball(r)| = \volcon_dr^d$.  Then, simplifying this expression 
yields the following Lemma:

\begin{lemma}[Probability of Path Coverage]
\label{lem:pcover}
Let \cover\ be the event that for one execution of \prmstar\, there 
exists at least one sample in each of the $\balls+1$ hyperballs 
of radius $\ballrad$ over the clearance robust optimal path, \piopt, 
for a specific value of $n > \nnaught$ and $\ballrad$.  Then,
\begin{equation}
\pcover \approx \bigg( 1 - \Big( 1 - a \Big)^n \bigg)^{ \frac{1}{2}
\lopt\big( b \big)^{-\frac{1}{d}} + 1 } 
\label{eq:pcover}
\end{equation}
Where $a = \Big( 1 + \frac{1}{d} \Big)\Big( \frac{\log{n}}{n} \Big)$ 
and $b = \Big( 1 + \frac{1}{d} \Big) \Big( \frac{|\cfree|\log{n}}
{\volcon_dn} \Big)$.
\end{lemma}

\subsection{Deriving a probabilistic bound}
\label{sec:chebyshev}

Let $\ccover$ be the event that there does not exist a sample in each
of the hyperballs covering a path, i.e. $\pr(\ccover) = 1 - 
\pcover$.  Then, the value for $\pr( |\lalg - \lopt| > \delta\cdot
\lopt )$ can be expressed as: 

\begin{multline*}
\pr\big( |\lalg -\lopt| > \delta\cdot\lopt\ |\ \cover \big)\pr( \cover )\\ + 
\pr\big( |\lalg -\lopt| > \delta\cdot\lopt\ |\ \ccover \big)\pr( \ccover )
\end{multline*}

\noindent This is because the probability of returning a low quality 
path is expressed as a sum of probabilities, when 
event \cover\ has occurred, and when \cover\ has not 
occurred.  Since $\pr (\ccover) = 1 - \pcover$, then via Lemma 
\ref{lem:pcover}, both $\pr(\ccover)$ and $\pcover$ are known for 
known $n$ and $\beta$.  It is assumed that the probability of a 
path being larger than $\delta$ is quite high if 
$\cover$ has not happened, i.e. $\pr\big( |\lalg -\lopt| > \delta\cdot
\lopt\ |\ \ccover \big)$ is close to $1$; therefore, this probability 
can be upper bounded by $1$.  
All that remains is to compute $\pr\big( |\lalg -\lopt| > \delta
\lopt\ |\ \cover \big)$.  Let $\zmu$ be a random variable
identically distributed with $\lalg$, but having $0$ mean, i.e. $\zmu = 
\lalg - \expect[\lalg]$.  Then, let

\begin{multline*}
\pr\big( |\lalg -\lopt| > \delta\lopt\ |\ \cover \big) = \\ \pr\big( 
|\expect[\lalg] + \zmu -\lopt| > \delta\lopt\ |\ \cover \big)
\end{multline*}

Then, the absolute value can be removed, as $|X| > a \Rightarrow 
X > a$ or $X<-a$.  Then, the probability is equal to the sum:

\begin{multline*}
\pr\big( \expect[\lalg] + \zmu -\lopt > \delta\lopt\ |\ \cover \big)\\ + 
\pr\big( \expect[\lalg] + \zmu -\lopt < -\delta\lopt\ |\ \cover \big) ,
\end{multline*}

\noindent where due to symmetry,

\begin{multline*}
\pr\big( |\lalg -\lopt| > \delta\lopt\ |\ \cover \big) = \\
2\pr\big( \expect[\lalg] + \zmu -\lopt > \delta\lopt\ |\ \cover \big)
\end{multline*}

\noindent Rearranging the terms inside the probability yields:

\begin{multline*}
\pr\big( |\lalg -\lopt| > \delta\lopt\ |\ \cover \big) = \\
2\pr\big( \zmu > (\delta+1)\lopt - \expect[\lalg]\ |\ \cover \big)
\end{multline*}

\noindent This probability will be bounded with Chebyshev's 
Inequality, which states:

$$\pr(|X-\expect[X]| \geq a)\leq \frac{Var(X)}{a^2} $$

\noindent In order to employ this inequality, both $\expect[\lalg]$ and
$Var(\lalg)$ for the length of a path in the \prmstar\ planning 
structure, \lalg\ are needed.

\subsection{Approximation of $\expect[\lalg]$ in $\reals^d$}

\begin{figure}[t]
\includegraphics[width=2.53in]{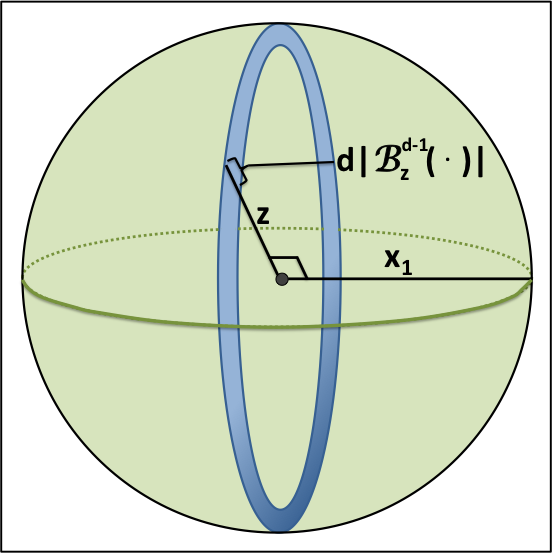}
\centering
\caption{The differential over a lower-dimensional hyperball, 
illustrated for $d=3$. }
\label{fig:differential}
\end{figure}

\begin{figure}
\centering
\includegraphics[width=2.7in]{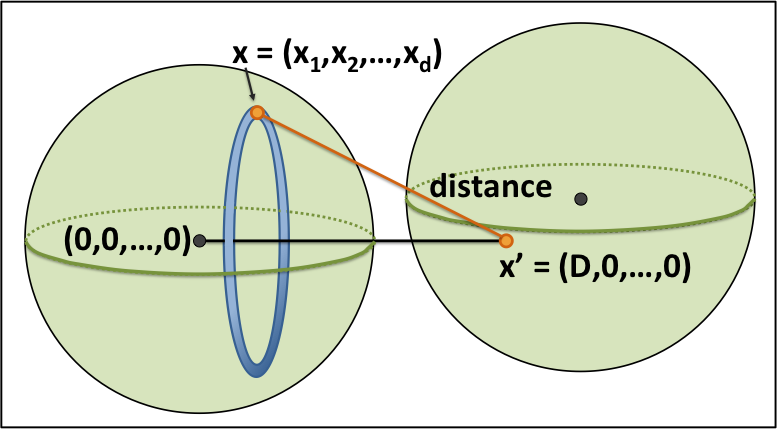}
\includegraphics[width=2.7in]{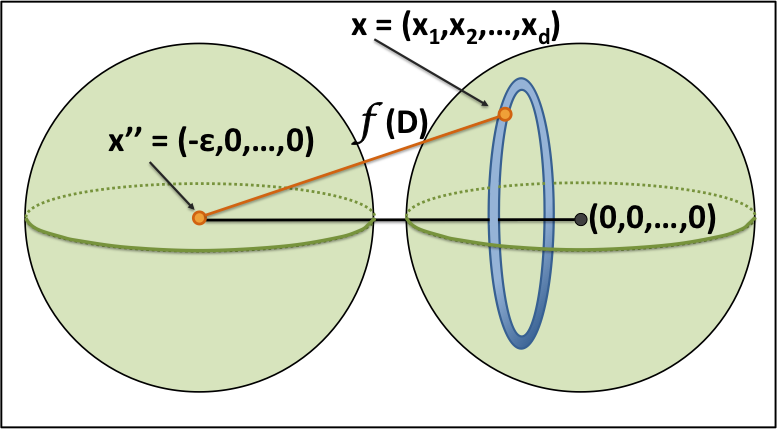}
\vspace{-0.1in}
\caption{Illustrations in 3D of the mean calculation.  (top) The 
first set of integrals is performed over the left hyperball, averaging
the distance between points $(x_1,x_2,\ldots,x_d)$ and $(D,0,
\ldots,0)$. (bottom) Using the result from the first set of integrals,
a second set of integrals is performed over the second hyperball,
yielding the expected value.}
\label{fig:mean_integration}
\end{figure}

Let, $\expect[\lalg] = \sum_{m=1}^M \expect[\lm]$, where $\lm$
is the length of a single segment between two random samples
in consecutive disjoint balls.  Then, because all $\lm$ are I.I.D.,
$\expect[\lalg] = M\expect[\lo]$.  Then, to compute $\expect[\lalg]$, 
$\expect[\lo]$ is computed.  This problem is similar to the problem 
known as the ball-line picking problem from geometric probability 
\cite{Santalo1976Integral-Geomet}.  The ball-line picking
problem is to compute the average length of a segment lying within 
a d-dimensional hyperball, where the endpoints of the segment are
uniformly distributed within the hyperball.  The ball-line picking
problem yields an analytical solution in general dimension; however, 
in the problem examined here, there are two disjoint hyperballs
rather than a single hyperball.  To the best of the authors' 
knowledge, this variant of the problem has not been previously 
studied.  Computing this value requires
integration over the possible locations of the endpoints of the 
segment, as illustrated in Figure \ref{fig:mean_integration}.  

The integration is broken into two steps, and the first integral
will be for the situation depicted in Figure \ref{fig:mean_integration}
(left).  The objective is to get an expected value for the distance 
between points $x$ and $x'$. Here, $x$ represents a random point 
within the first hyperball, while $x'$ is some fixed point within the 
second hyperball which has distance $D$ from the center of the first 
hyperball.  Without loss of generality, $x'$ can be displaced along 
only the first coordinate, $x_1$.  To get an expected value, this 
distance is integrated over all points within the first hyperball, 
and then divided by the volume of the d-dimensional hyperball.  
In this work, the volume of a d-dimensional hypersphere of radius 
$\ballrad$ is denoted $|\ball_{\ballrad}| = \volcon_d\ballrad^d$, 
where $\volcon_d$ is a constant dependent on the dimension of the space. 
Taking the distance between $x = (x_1,x_2,\ldots, x_d)$ and $x' = 
(D,0,\ldots,0)$ to be $\sqrt{(x_1 - D)^2 + x_2^2 + \ldots + x_d^2}$ 
produces the following integral:

\begin{multline*}
A = \frac{1}{\volcon_d\ballrad^d} \idotsint_{x_1^2 + \ldots + x_d^2 \leq 
\ballrad^2} \\ \sqrt{(x_1 - D)^2 + x_2^2 + \ldots + x_d^2}\ dx_1 \ldots dx_d, 
\end{multline*}

This integral will be converted from a d-dimensional integral into
a double integral using substitution.  First, let $z^2 = 
x_2^2 + \ldots + x_d^2$.  This allows performing the integral over 
only two variables, $x_1$ and $z$; however, the form of the integral
changes, as the differential is adapted as illustrated in Figure
\ref{fig:differential}.  This differential, $d|\ball_{d-1}(z)|$, is taken
over a lower dimensional hypersphere, of dimension $d-1$, as $z$ is
taking the place of $d-1$ coordinates. Then:

\begin{multline*}
A = \frac{1}{\volcon_d\ballrad^d} \iint_{x_1^2 + z^2 \leq \ballrad^2} \\
\sqrt{(x_1 - D)^2 + z^2}\ d|\ball^{d-1}_{z}(\cdot)| dx_1 dz ,
\end{multline*}

\noindent where $d|\ball_{d-1}(z)| = \frac{d}{dz}\volcon_{d-1}z^{d-1}$.  
Taking this derivative, $ \frac{d}{dz}\volcon_{d-1}z^{d-1} = (d-1)
\volcon_{d-1}z^{d-2}$, and substituting into $A$ yields:

\begin{multline*}
A = \frac{(d-1)\volcon_{d-1}}{\volcon_d\ballrad^d} \iint_{x_1^2 + z^2
\leq \ballrad^2} \\ z^{d-2} \sqrt{x_1^2 + D^2 - 2Dx_1 + z^2}\ dx_1 dz 
\end{multline*}

The integral can be represented in terms of polar coordinates, where 
$x_1 = r \cos\theta$, $z = r \sin\theta$, and $dx_1\ dz = r\ 
d\theta\ dr$.  This gives

\begin{multline*}
A = \frac{(d-1)\volcon_{d-1}}{\volcon_d\ballrad^d} \int_0^{\ballrad} r
\int_0^\pi \\ (r \sin\theta)^{d-2} \sqrt{r^2 + D^2 - 
2Dr\cos\theta}\ d\theta dr 
\end{multline*}

\begin{multline*}
A = \frac{D(d-1)\volcon_{d-1}}{\volcon_d\ballrad^d} \int_0^{\ballrad} r
\int_0^\pi \\ (r \sin\theta)^{d-2} \sqrt{1 + \big(\frac{r}{D}\big)^2 - 
2\big(\frac{r}{D}\big)\cos\theta}\ d\theta dr 
\end{multline*}

\noindent A second-order Taylor Approximation for the square root is 
taken.  Let $f(u) = \sqrt{1 + u}$, where $u = \big(\frac{r}{D}\big)^2
- 2\big(\frac{r}{D}\big)\cos\theta$.  The approximation will be 
taken about the point $u = 0$.  This is reasonable given that overall, 
$\ballrad$ is considered to be smaller than the separation between 
consecutive hyperballs, $\clear$. Take the second-order Taylor 
Approximation as:

$$f(u) \approx f(0) + f'(0)\cdot u + \frac{1}{2!} f''(0)\cdot u^2 .$$

\noindent Taking a derivative of $f$ yields $f'(u) = \frac{1}{2}
(1+u)^{-\frac{1}{2}}$ and $f''(u) = -\frac{1}{4}(1+u)^{-\frac{3}{2}}$.
Then,

$$f(u) \approx \sqrt{1} + \frac{1}{2}(1)^{-\frac{1}{2}}\cdot u 
-\frac{1}{2}\cdot\frac{1}{4}(1)^{-\frac{3}{2}} \cdot u^2 = 1 + 
\frac{1}{2}u -\frac{1}{8} u^2$$

\noindent substituting $u = \big(\frac{r}{D}\big)^2 - 2\big(
\frac{r}{D}\big)\cos\theta$,

\begin{multline*}
f(u) \approx 1 + \frac{1}{2}\big( \big(\frac{r}{D}\big)^2 
- 2\big(\frac{r}{D}\big)\cos\theta \big)\\ - \frac{1}{8}\big(
\big(\frac{r}{D}\big)^2 - 2\big(\frac{r}{D}\big)\cos\theta \big)^2 
\end{multline*}

\begin{multline*}
f(u) \approx 1 + \frac{1}{2}\big( \big(\frac{r}{D}\big)^2 
- 2\big(\frac{r}{D}\big)\cos\theta \big)\\ - \frac{1}{8}\big( 
\big(\frac{r}{D}\big)^4 -4\big(\frac{r}{D}\big)^3\cos\theta
+ 4\big(\frac{r}{D}\big)^2\cos^2\theta \big)
\end{multline*}

\noindent Then, as this is a second-order approximation, the third-
and fourth-order terms are considered negligible, and thus, the
approximation results in:

$$ f(u) \approx 1 + \frac{1}{2}\big( \big(\frac{r}{D}\big)^2 
- 2\big(\frac{r}{D}\big)\cos\theta \big) - \frac{1}{8}\big( 
4\big(\frac{r}{D}\big)^2\cos^2\theta \big) $$

Substituting the result:

\begin{multline*}
A = \frac{D(d-1)\volcon_{d-1}}{\volcon_d\ballrad^d} \int_0^{\ballrad} r
\int_0^\pi (r \sin\theta)^{d-2} \\ \Big( 1 + \frac{1}{2}\big( 
\big(\frac{r}{D}\big)^2 + 2\big(\frac{r}{D}\big)\cos\theta\big) -\frac{1}{8}
\big(4\big(\frac{r}{D}\big)^2\cos^2\theta\big) \Big) \ d\theta dr 
\end{multline*}

Simplifying this integral requires the following Lemmas:

\begin{lemma}[Value of $\int_0^\pi (\sin\theta)^d d\theta$]
In terms of the hyperball volume constant, $\volcon_d$, 
$$ \int_0^\pi (\sin\theta)^{d-2} d\theta =  \sint_{d-2} = 
\frac{d\volcon_d}{(d-1)\volcon_{d-1}} $$
\label{lem:sine_volume}
\end{lemma}

\noindent{\bf Proof}

For simplicity, let $\int_0^\pi sin^d(\theta)\ d\theta$ be denoted
as $\sint_d$.  Then:

$$ |\ball_\beta| = \volcon_d \beta^d ,$$

where $\volcon_d = \frac{\pi^{\frac{d}{2}}}{\Gamma(\frac{d}{2}+1)}$
is a constant dependent on the dimension, $d$.  Then, the volume can
be computed as an integral of the following form:

$$ \volcon_d \beta^d = \iint_{x_1^2 + \rho^2 \leq \beta^2}\ dx_1\ 
d(\volcon_{d-1}\rho^{d-1}) ,$$

where the second differential is over a sphere of radius $\rho$ of
dimension $d-1$.  Then,

$$ \volcon_d \beta^d = \iint_{x_1^2 + \rho^2 \leq \beta^2} (d-1)
\volcon_{d-1} \rho^{d-2} \ dx_1\ d\rho$$

Now, to simplify this integral, it will be converted to polar 
coordinates, using $x_1 = r\ \cos\theta$, $\rho = r\ \sin\theta$,
and $dx_1\ d\rho = r\ d\theta\ dr$.  Substituting these values yields:

$$ \volcon_d \beta^d = (d-1) \volcon_{d-1} \int_0^\beta \int_0^\pi 
r^{d-2} (\sin\theta)^{d-2} r\ d\theta\ dr$$

$$ \volcon_d \beta^d = (d-1) \volcon_{d-1} \int_0^\beta  
r^{d-1} \sint_{d-2}\ d\theta\ dr$$

$$ \volcon_d \beta^d = (d-1) \volcon_{d-1} \sint_{d-2} \frac{\beta^d}{d}
\ d\theta\ dr$$

$$ \sint_{d-2} = \frac{d\volcon_d}{(d-1)\volcon_{d-1}}\ \square$$

\begin{lemma}[Recurrence relation of $\int_0^\pi (\sin\theta)^d d\theta$]
For $\sint_d = \int_0^\pi (\sin\theta)^d d\theta$, the following
recurrence relation holds:
$$ \int_0^\pi (\sin\theta)^d d\theta = \sint_d = \frac{d-1}{d}\sint_{d-2} $$
\label{lem:recurrence}
\end{lemma}

\noindent{\bf Proof}
To determine this recurrence, the following expression will be solved
for $x$:

$$ \sint_d = x \sint_{d-2}$$

Substitute the result from Lemma \ref{lem:sine_volume}, getting:

$$ x = \frac{(d+2)\volcon_{d+2}}{(d+1)\volcon_{d+1}}
\cdot \frac{(d-1)\volcon_{d-1}}{d\volcon_d} .$$

Then, substituting the value of $\volcon_d$ yields:


$$ x = \frac{(d-1)(d+2)}{d(d+1)}
\Big( \frac{\Gamma(\frac{d+2}{2})}{\Gamma(\frac{d+4}{2})} \Big) 
\Big( \frac{\Gamma(\frac{d+3}{2})}{\Gamma(\frac{d+1}{2})} \Big)$$

$$ x = \frac{(d-1)(d+2)}{d(d+1)}
\Big( \frac{2}{d+2} \Big) 
\Big( \frac{d+1}{2} \Big)$$

$$ x = \frac{d-1}{d}\ \square $$

\noindent Applying these Lemmas:

\begin{multline*}
A = \frac{D(d-1)\volcon_{d-1}}{\volcon_d\beta^d} \\  \int_0^{\ballrad} r
\Bigg( \int_0^\pi (r \sin\theta)^{d-2} \Big( 1 + \frac{1}{2} \big(
\frac{r}{D}\big)^2 \Big) d\theta \\+ \int_0^\pi (r \sin\theta)^{d-2}
(\cos\theta) d\theta\\ -\frac{1}{2}\big(\frac{r}{D}\big)^2 \int_0^\pi 
(sin\theta)^{d-2}cos^2\theta\ d\theta \Bigg) dr 
\end{multline*}

The second integral over $\theta$ will integrate to $0$, due to the
presence of cosine, while the other terms leverage Lemmas
\ref{lem:sine_volume} and \ref{lem:recurrence}:

\begin{multline*}
A = \frac{D(d-1)\volcon_{d-1}}{\volcon_d\ballrad^d} \int_0^{\ballrad} \\ 
r^{d-1} \Bigg( \sint_{d-2} \Big( 1 + \frac{1}{2} \big(\frac{r}{D}\big)^2 
\Big) -\frac{1}{2}\big(\frac{r}{D}\big)^2 \Big( \sint_{d-2} - \sint_d 
\Big) \Bigg) dr
\end{multline*}

\begin{multline*}
A = \frac{D(d-1)\volcon_{d-1}}{\volcon_d\ballrad^d} \sint_{d-2} \cdot
\\ \int_0^{\ballrad} r^{d-1} \Bigg( \Big( 1 + \frac{d-1}{2d} \big(
\frac{r}{D}\big)^2 \Big) \Bigg) dr
\end{multline*}

\begin{multline*}
A = \frac{D(d-1)\volcon_{d-1}}{\volcon_d\ballrad^d} \sint_{d-2} \cdot
\\ \int_0^{\ballrad} \Bigg( r^{d-1} + \frac{d-1}{2d} \Big(
\frac{r^{d+1}}{D^2}\Big) \Bigg)dr 
\end{multline*}

\begin{multline*}
A = \frac{D(d-1)\volcon_{d-1}}{\volcon_d\ballrad^d} \frac{d\volcon_d}
{(d-1)\volcon_{d-1}} \cdot \\ \Bigg( \frac{\ballrad^{d}}{d} + 
\frac{d-1}{2d} \Big(\frac{\ballrad^{d+2}}{(d+2)D^2}\Big) \Bigg) \\ = 
D + \frac{(d-1)\ballrad^2}{2(d+2)D} 
\end{multline*}

This is only an intermediate result, however, and it must be 
integrated over once again to consider all possible placements of 
the point $x'$ in the second hyperball, as illustrated in Figure 
\ref{fig:mean_integration}(right).  In order to do so, write $D$ in 
terms of $\epsilon$ by taking the distance between $x = (x_1, x_2, 
\ldots, x_d)$ and $x'' = (-\clear,0,\ldots,0)$.  Then, $D = 
\sqrt{(x_1 + \epsilon)^2 + x_2^2 + \ldots + x_d^2}$, and 
$\expect[\lo]$ is computed as:

\begin{multline*}
\expect[\lo] = \frac{1}{\volcon_d\ballrad^d} \idotsint_{x_1^2 + 
\ldots + x_d^2 \leq \ballrad^2} D \\ + \frac{(d-1)\ballrad^2}{2(d+2)D}\ dx_1 
\ldots dx_d, 
\end{multline*}

Steps similar to what was just taken to derive the intermediate 
result are used to compute this integral.
As a matter of simplicity, note that the second term inside the
integral is already a second-order term, which means taking the
integral will result in higher-order terms.  Since $D = 
\sqrt{(x_1 + \clear)^2 + x_2^2 + \ldots + x_d^2}$, the second
term will take only the constant term of the Taylor Approximation
for $D$.  Then, taking $z^2 = x_2^2 + \ldots + x_d^2$:

\begin{multline*}
\expect[\lm] = \frac{1}{\volcon_d\ballrad^d} \idotsint_{x_1^2 + 
\ldots + x_d^2 \leq \ballrad^2}\\ \sqrt{(x_1 + \clear)^2 + x_2^2 + 
\ldots + x_d^2}\ \\+ \frac{(d-1)\ballrad^2}{2(d+2)\sqrt{(x_1 + \clear)^2 
+ x_2^2 + \ldots + x_d^2}}\ dx_1 \ldots dx_d 
\end{multline*}

\begin{multline*}
\expect[\lm] = \frac{1}{\volcon_d\ballrad^d} \iint_{x_1^2 + 
z^2 \leq \ballrad^2} \sqrt{(x_1 + \clear)^2 + z^2}\\ + \frac{(d-1)\ballrad^2}
{2(d+2)\sqrt{(x_1 + \clear)^2 
+ z^2}}\ dx_1\ dz 
\end{multline*}

Again, perform a polar coordinate transformation so as to take the
integral:

\begin{multline*}
\expect[\lm] = \frac{1}{\volcon_d\ballrad^d} \int_0^{\ballrad} r \int_0^\pi 
\Big( \sqrt{r^2 + \clear^2 - 2\clear r\cos\theta} \\+
\frac{(d-1)\ballrad^2}{2(d+2)\sqrt{r^2 + \clear^2 - 2\clear 
r\cos\theta}}\Big)d|\ball^{d-1}_{r \sin\theta}(\cdot)|\ d\theta\ dr 
\end{multline*}

\begin{multline*}
\expect[\lm] = \frac{(d-1)\volcon_{d-1}}{\volcon_d\ballrad^d} 
\int_0^{\ballrad} r \int_0^\pi (r\ \sin\theta)^{d-2}\\ \Big( \sqrt{r^2 + 
\clear^2 - 2\clear r\cos\theta} \\+ \frac{(d-1)\ballrad^2}{2(d+2)
\sqrt{r^2 + \clear^2 - 2\clear r\cos\theta}}\Big)\ d\theta\ dr
\end{multline*}

Now, to compute this integral, a Taylor Approximation will be taken.
Again, recall that because the second term in the integral is
second order, a $0^{th}$-order approximation is taken for that term:

\begin{multline*}
\expect[\lm] \approx \frac{(d-1)\volcon_{d-1}}{\volcon_d\ballrad^d} 
\int_0^{\ballrad} r \int_0^\pi (r\ \sin\theta)^{d-2}\\ \Big(\clear\Big( 1 + 
\frac{1}{2}\big( \big(\frac{r}{\clear}\big)^2 + 2\big(\frac{r}
{\clear}\big)\cos\theta\big) -\frac{1}{8}\big(4\big(\frac{r}{\clear}
\big)^2\cos^2\theta\big) \Big)\\ + \frac{(d-1)\ballrad^2}{2(d+2)\clear}
\Big)\ d\theta\ dr 
\end{multline*}

Then, performing steps similar to above, rewrite in terms of 
$\sint_{d-2}$, as well as splitting the last term into a separate
integral:

\begin{multline*}
\expect[\lm] \approx \frac{\clear(d-1)\volcon_{d-1}}{\volcon_d\ballrad^d} 
\int_0^{\ballrad} r^{d-1} \int_0^\pi \Big( (\sin\theta)^{d-2}\\ + 
\frac{r^2}{2\clear^2}(\sin\theta)^d \Big)\ d\theta\ dr + 
\frac{(d-1)\volcon_{d-1}}{\volcon_d\ballrad^d} \cdot \\ \int_0^{\ballrad} r^{d-1} 
\int_0^\pi (\sin\theta)^{d-2} \frac{(d-1)\ballrad^2}{2(d+2)\clear}
\ d\theta\ dr
\end{multline*}

\begin{multline*}
\expect[\lm] \approx \frac{\clear(d-1)\volcon_{d-1}}{\volcon_d\ballrad^d} 
\sint_{d-2} \int_0^{\ballrad} r^{d-1} \Big( 1\\ + \frac{(d-1)r^2}{2d\clear^2}
\Big)\ dr + \frac{\clear(d-1)\volcon_{d-1}}{\volcon_d\ballrad^d} 
\sint_{d-2} \\ \int_0^{\ballrad} r^{d-1} \frac{(d-1)\ballrad^2}{2(d+2)
\clear^2}\ dr 
\end{multline*}

\begin{multline*}
\expect[\lm] \approx \frac{\clear(d-1)\volcon_{d-1}}{\volcon_d\ballrad^d} 
\sint_{d-2} \Big( \frac{\ballrad^d}{d} + \frac{(d-1)\ballrad^{d+2}}{2d(d+2)
\clear^2}\Big)\\ + \frac{\clear(d-1)\volcon_{d-1}}{\volcon_d\ballrad^d} 
\sint_{d-2} \Big( \frac{(d-1)\ballrad^{d+2}}{2d(d+2)\clear^2} \Big) 
\end{multline*}

\begin{multline*}
\expect[\lm] \approx \frac{\clear(d-1)\volcon_{d-1}}{\volcon_d\ballrad^d} 
\frac{d\volcon_{d-1}}{(d-1)\volcon_{d-1}}\cdot \\ \Big( \frac{\ballrad^d}{d} + 
\frac{(d-1)\ballrad^{d+2}}{2d(d+2)\clear^2} + \frac{(d-1)\ballrad^{d+2}}
{2d(d+2)\clear^2} \Big)
\end{multline*}

$$\expect[\lm] \approx \clear + \frac{(d-1)\ballrad^2}{(d+2)\clear} $$

Now, using this result for the expected value of a single segment,
$\lm$, the expected value of the entire path consisting of $M$ such
segments is:

$$ \expect[\lalg] \approx M\Big(\clear + \frac{(d-1)\ballrad^2}{(d+2)
\clear} \Big) $$

\begin{lemma}[Expected value of $\lalg$]
\label{lem:mean}
For a path constructed over the set of $\balls + 1$ hyperballs 
having radius $\ballrad$ has expected length:
$$\expect[\lalg] \approx \balls\Big(\clear + \frac{(d-1)\ballrad^2}{(d+2)
\clear} \Big) $$
\end{lemma}

\begin{figure}
  \centering
  \begin{tabular}{| c | c | c |}
    \hline
    Euclidean & $\lan = 0.5$ & $\lan = 0.125$ \\
    \hline
    dimension & $100\cdot\frac{|\expect[\lalg] - \lalg|}{\lalg}$ & 
    $100\cdot\frac{|\expect[\lalg] - \lalg|}{\lalg}$ \\
    \hline
    2 & 0.1730\% & 0.0050\% \\
    3 & 0.0473\% & 0.0205\% \\
    10 & 0.9413\% & 0.0128\% \\
    100 & 1.9147\% & 0.0129\% \\
    \hline
  \end{tabular}
  \vspace{-0.13in}
  \caption{Simulation comparison for $\expect[\lalg]$, using 120,000 
  data points for each entry, for differing $\lan = \frac{\ballrad}
  {\clear}$}
  \label{tab:mean_tight}
\end{figure}
To verify that the approximation for the expected value is tight,
Monte Carlo experiments were run and compared against the drawn
approximate value.  The relative error of the approximation to the
simulated values are shown in Figure \ref{tab:mean_tight}.  The 
computed approximation deviates more from experimental data as
dimensionality increases, but holds quite well, especially for small
values of $\lan = \frac{\ballrad}{\clear}$.

\subsection{Computation of the Variance of $\lalg$ in $\reals^d$}

To compute the $Var(\lalg)$, leverage the definition of the
variance of a random variable, i.e. $ Var(X) = \expect[X^2] - 
(\expect[X])^2 $:

\begin{multline*}
Var\big( \sum_{m=1}^{M} \lm \big) = \expect[\sum_{m=1}^{M} \lm^2] 
- \big(\expect[\sum_{m=1}^{M} \lm] \big)^2 \\ = \sum_{m=1}^{M}
\sum_{k=1}^{M} \expect[\lm \lkk] - \big(\expect[\sum_{m=1}^{M} \lm] 
\big)^2
\end{multline*}

The second term can be simplified due to the linearity of expectation,
which allows the double sum to be simplified:

$$ Var\big( \sum_{m=1}^{M} \lm \big) = \sum_{m=1}^{M}\sum_{k=1}^{M}
\expect[\lm \lkk] - M^2\big(\expect[\lm] \big)^2 $$

Then, the first term consists of the expected value of the product of
all $M^2$ pairs of segments along a path.  There are variance terms 
for each segment with itself, i.e. $\expect[\lm^2]$, of which there 
are $M$.  Additionally, pairs of segments which share an endpoint 
have a dependence.  Only consecutive pairs have such a dependence, 
and each segment depends on two neighbors, or a total of $2M$ 
dependencies, except that the start and end segments only have a 
single neighbor, which yields a total of $2M-2$ such dependencies.  
Then, all of the other segments must be independent.  Then, expand 
the double sum as:

\begin{multline*}
\sum_{m=1}^{M}\sum_{k=1}^{M}\expect[\lm \lkk] = M\expect[\lm^2] + 
(2M-2)\expect[\lm\lk] \\+ (M^2 - M - (2M-2))\expect[\lm\lj] 
\end{multline*}

where $\lm$ and $\lj$ are independent, and $\lm$ and $\lk$ are 
dependent consecutive segments.  Due to this independence, and
substituting $m=1$ yields:

\begin{multline*}
Var\big( \sum_{m=1}^{M} \lm \big) = M\expect[\lo^2] + (2M-2)
\expect[\lo\lt] \\+ (M^2 - M - (2M-2))\expect[\lo]\expect[\lo] - 
M^2\big(\expect[\lo] \big)^2 
\end{multline*}

This results in a final variance term:

\begin{multline*}
Var\big( \sum_{m=1}^{M} \lm \big) = M\expect[\lo^2] \\ + (2M-2)
\expect[\lo\lt] + (2 - 3M)\big(\expect[\lo]\big)^2 
\end{multline*}

From the previous section, the value for $(\expect[\lo])^2$ is 
available; however, both $\expect[\lo^2]$ and $\expect[\lo\lt]$ 
must be computed.

\subsubsection{Approximation of $\expect[\lo^2]$ in $\reals^d$.}  

The derivation of $\expect[\lo^2]$ follows the same general steps as 
the computation of $\expect[\lo]$; however, the form of the integral 
is simpler in this case.  The integral over one of the two balls is 
of the form:

\begin{multline*}
A = \frac{1}{\volcon_d{\ballrad}^d}\idotsint_{x_1^2 + \ldots + x_d^2 
\leq {\ballrad}^2}\\ \Big(\sqrt{(D-x_1)^2 + x_2^2 + \ldots + x_d^2}\Big)^2 
dx_1 \dots dx_d ,
\end{multline*}

$A$ represents an intermediate result.  Then, take
$z^2 = x_2^2 + \ldots + x_d^2$.  Substituting these into the above
integral yields:

$$\frac{1}{\volcon_d{\ballrad}^d}\iint_{x_1^2 + z^2 \leq {\ballrad}^2}
\big((D-x_1)^2 + z^2\ d|\ball^{d-1}_z(\cdot)|\big)\ dx_1 dz $$

\begin{multline*}
= \frac{1}{\volcon_d{\ballrad}^d}\iint_{x_1^2 + z^2 \leq {\ballrad}^2}
\\ \volcon_{d-1}(d-1)z^{d-2} \big((D-x_1)^2 + z^2\big)\ dx_1 dz 
\end{multline*}

Then, to compute this integral, perform the integration over
polar coordinates:

\begin{multline*}
\frac{\volcon_{d-1}(d-1)}{\volcon_d{\ballrad}^d} \int_{0}^{\ballrad} r \int_{0}^\pi
(r\ \sin\theta)^{d-2}\\ \big((D-(r\ \cos\theta))^2 + (r\ \sin\theta)^2
\big)\ d\theta\ dr 
\end{multline*}

\begin{multline*}
= \frac{\volcon_{d-1}(d-1)}{\volcon_d{\ballrad}^d} \int_{0}^{\ballrad}
r^{d-1} \int_{0}^\pi (\sin\theta)^{d-2}\\ \big( D^2 + r^2 - 2Dr\cos\theta 
\big)\ d\theta\ dr 
\end{multline*}

Then, applying Lemmas \ref{lem:sine_volume} and \ref{lem:recurrence}:

$$\frac{\volcon_{d-1}(d-1)}{\volcon_d{\ballrad}^d} \int_{0}^{\ballrad} r^{d-1} 
\big( D^2 + r^2 \big)\sint_{d-2}\ d\theta\ dr $$

$$= \frac{\volcon_{d-1}(d-1)\sint_{d-2}}{\volcon_d{\ballrad}^d} \big( 
D^2\frac{{\ballrad}^d}{d} + \frac{d{\ballrad}^{d+2}}{d(d+2)} \big)\ d\theta\ dr $$

$$A = D^2 + \frac{d}{d+2}{\ballrad}^2 $$

Now, this intermediate result is used to compute the final value of 
$\expect[\lo^2]$.  Begin again by integrating over this value:

\begin{multline*}
\expect[\lo^2] = \frac{1}{\volcon_d{\ballrad}^d}\idotsint_{x_1^2 + 
\ldots + x_d^2 \leq {\ballrad}^2} D^2\\ + \frac{d}{d+2}{\ballrad}^2 dx_1 
\dots dx_d ,
\end{multline*}

where now $D$ is written in terms of $\clear$ as $D = \sqrt{
(x_1 + \clear)^2 + x_2^2 + \ldots + x_d^2}$.  Again, take 
$z^2 = x_2^2 + \ldots + x_d^2$ and substitute in to get:

\begin{multline*}
\frac{1}{\volcon_d{\ballrad}^d}\iint_{x_1^2 + z^2 \leq {\ballrad}^2}
\\ \big(((x_1 + \clear)^2 + z^2) + \frac{d}{d+2}{\ballrad}^2\big) 
d|\ball^{d-1}_z(\cdot)|\ dx_1\ dz
\end{multline*}

\begin{multline*}
= \frac{(d-1)\volcon_{d-1}}{\volcon_d{\ballrad}^d}\iint_{x_1^2 + z^2 
\leq {\ballrad}^2} \\ \big((r^2 + \clear^2 + 2x_1\clear) + \frac{d}
{d+2}{\ballrad}^2\big) z^{d-2}\ dx_1\ dz 
\end{multline*}

Rewrite in polar coordinates and splitting the integral:

\begin{multline*}
\frac{(d-1)\volcon_{d-1}}{\volcon_d{\ballrad}^d} \Bigg( \int_0^{\ballrad} r 
\int_0^\pi r^d (\sin\theta)^{d-2}\\ + (r \sin\theta)^{d-2}\clear^2 
+ 2\clear r\cos\theta d\theta\ dr \\+ \frac{d}{d+2}{\ballrad}^2 \int_0^{\ballrad} 
r \int_0^\pi (r \sin\theta)^{d-2}\ d\theta\ dr \Bigg) 
\end{multline*}

\begin{multline*}
= \frac{(d-1)\volcon_{d-1}}{\volcon_d{\ballrad}^d} \Bigg( \int_0^{\ballrad} r
\big( \sint_{d-2}(r^d + \clear^2r^{d-2}) \big) dr\\ + \frac{d}{d+2}
{\ballrad}^2 \int_0^{\ballrad} r^{d-1}\sint_{d-2} dr \Bigg) 
\end{multline*}

$$= \frac{(d-1)\volcon_{d-1}}{\volcon_d{\ballrad}^d} \sint_{d-2} \Bigg(
\frac{{\ballrad}^{d+2}}{d+2} + \frac{\clear^2{\ballrad}^d}{d} + 
\frac{d{\ballrad}^2}{d+2} + \frac{{\ballrad}^d}{d} \Bigg) $$

$$\expect[\lo^2] = \clear^2 + \frac{2d}{d+2}{\ballrad}^2 $$

\begin{lemma}[Expected value of $\lo^2$]
For two consecutive hyperballs, the expected squared distance
between random points in those spheres is
$$\expect[\lo^2] = \clear^2 + \frac{2d}{d+2}\ballrad^2 $$
\end{lemma}

\subsubsection{Approximation of $\expect[\lo\lt]$ in $\reals^d$.}

\begin{figure}
\centering
\includegraphics[width=2.74in]{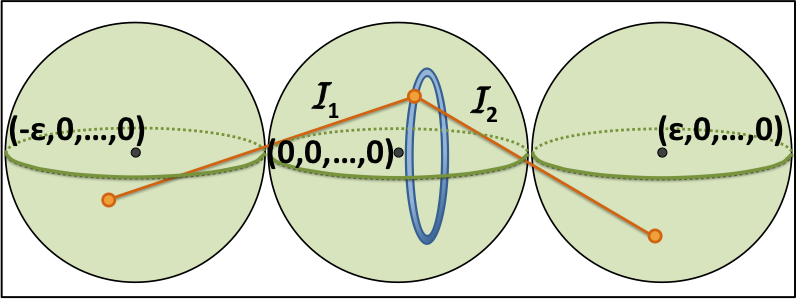}
\caption{To compute $\expect[\lo\lt]$, integration is performed over
the common point determining $\lo$ and $\lt$.}
\label{fig:variance}
\end{figure}
To compute $\expect[\lo\lt]$, a key observation is made.  First, to
retrieve this value, the reasoning must consider three consecutive
hyperballs, where the distance between the samples of the first two
balls, $\lo$, and the distance between the samples of the second and
third balls, $\lt$, depend on each other through their common endpoint
in the second ball.  Consider, however, that if this second point is
fixed, then the values of $\lo$ and $\lt$ become independent.  Using
this fact, and the intermediate result of the mean calculation, begin
by simply multiplying these two means to get the intermediate result:

$$A = (D_1 + \frac{C_d\ballrad^2}{2D_1})(D_2 + \frac{C_d\ballrad^2}{2D_2}) $$

$$A = D_1D_2 + \frac{C_d\ballrad^2}{2}\Big( \frac{D_1}{D_2} + \frac{D_2}{D_1} \Big) $$

\noindent where the fourth-order term involving $\ballrad^4$ is negligible.

Now, to reintroduce the dependence of $\lo$ and $\lt$, integrate over
the second ball.  For technical reasons, it is assumed that the centers
of the three hyperballs are collinear.  Then, the integral to solve
is the following: 

\begin{multline*}
\expect[\lo\lt] = \frac{1}{\volcon_d{\ballrad}^d} \idotsint_{x_1^2 
+ \ldots + x_d^2 \leq {\ballrad}^2} D_1D_2 \\ + \frac{(d-1)
{\ballrad}^2}{2(d+2)}\Big( \frac{D_2}{D_1} + \frac{D_1}{D_2} \Big)
\ dx_1 \ldots dx_d 
\end{multline*}

Where $D_1$ and $D_2$ are expressed in 
relative to the center of the middle hyperball as $D_1 = \sqrt{
(x_1 - \epsilon)^2 + x_2^2 + \ldots + x_d^2 }$ and $D_2 = \sqrt{
(x_1 + \epsilon)^2 + x_2^2 + \ldots + x_d^2 }$.  Again, taking
$z^2 = x_2^2 + \ldots x_d^2$ and then, because the second term 
is already second-order, take a $0^{th}$-order approximation over
only that term:

\begin{multline*}
\expect[\lo\lt] \approx \frac{1}{\volcon_d{\ballrad}^d} \iint_{x_1^2 
+ z^2 \leq {\ballrad}^2} \\
\Bigg( \sqrt{(x_1 - \clear)^2 + z^2}\sqrt{(x_1 + \clear)^2 + z^2}
\\+ \frac{(d-1){\ballrad}^2}{2(d+2)}\Big( \frac{\clear}{\clear} +
\frac{\clear}{\clear} \Big) \Bigg) d|\ball^{d-1}_z(\cdot)|\ dx_1 dz
\end{multline*}

Again, convert this integral into polar coordinates:

\begin{multline*}
\expect[\lo\lt] \approx \frac{(d-1)\volcon_{d-1}}{\volcon_d
{\ballrad}^d} \int_0^{\ballrad} r 
\int_0^\pi (r\ \sin\theta)^{d-2}\\ \Bigg( \sqrt{r^2 + \clear^2 -
2\clear r\cos\theta}\sqrt{r^2 + \clear^2 + 2\clear r\cos\theta}
\\+ \frac{(d-1){\ballrad}^2}{(d+2)} \Bigg)\ d\theta\ dr
\end{multline*}

\begin{multline*}
\expect[\lo\lt] \approx \frac{\clear^2(d-1)\volcon_{d-1}}{\volcon_d
{\ballrad}^d} \int_0^{\ballrad} 
r^{d-1} \int_0^\pi (\sin\theta)^{d-2}\\ \Bigg( \sqrt{1 + 2\Big( \frac{r}
{\clear} \Big)^2 - 4 \Big( \frac{r}{\clear} \Big)^2 \cos^2\theta }
+ \frac{(d-1){\ballrad}^2}{(d+2)} \Bigg)\ d\theta\ dr
\end{multline*}

Then, as the square root is prohibitive to integrate directly, a 
second-order Taylor approximation is employed:

\begin{multline*}
\expect[\lo\lt] \approx \frac{\clear^2(d-1)\volcon_{d-1}}{\volcon_d
{\ballrad}^d} \int_0^{\ballrad} r^{d-1} \int_0^\pi (\sin\theta)^{d-2}\\ 
\Bigg( 1 + \Big( \frac{r}
{\clear} \Big)^2 - 2\Big( \frac{r}{\clear} \Big)^2\cos^2\theta
+ \frac{(d-1){\ballrad}^2}{(d+2)} \Bigg)\ d\theta\ dr
\end{multline*}

\begin{multline*}
\expect[\lo\lt] \approx \frac{\clear^2(d-1)\volcon_{d-1}}{\volcon_d
{\ballrad}^d} \int_0^{\ballrad} r^{d-1} \int_0^\pi (\sin\theta)^{d-2}\\ 
\Bigg( 1 - \Big( \frac{r}
{\clear} \Big)^2 + 2\Big( \frac{r}{\clear} \Big)^2\sin^2\theta
+ \frac{(d-1){\ballrad}^2}{(d+2)} \Bigg)\ d\theta\ dr
\end{multline*}

\begin{multline*}
\expect[\lo\lt] \approx \frac{\clear^2(d-1)\volcon_{d-1}}{\volcon_d
{\ballrad}^d} \sint_{d-2} \int_0^{\ballrad} r^{d-1}\\ \Bigg( 1 - \Big( 
\frac{r} {\clear} \Big)^2 + \frac{2(d-1)}{d}\Big( \frac{r}{\clear} 
\Big)^2 + \frac{(d-1){\ballrad}^2}{(d+2)} \Bigg)\ dr
\end{multline*}

\begin{multline*}
\expect[\lo\lt] \approx \frac{d\clear^2}{{\ballrad}^d} \Bigg( \frac{{\ballrad}^d}{d} - 
\frac{{\ballrad}^{d+2}}{\clear^2(d+2)}\\ + \frac{(2d-2){\ballrad}^{d+2}}
{d\clear^2(d+2)} + \frac{(d-1){\ballrad}^2}{(d+2)} \Bigg) 
\end{multline*}

$$\expect[\lo\lt] \approx \clear^2 + \Big( \frac{2d-3}{d+2} \Big){\ballrad}^2 $$

\begin{lemma}[Expected value of $\lo\lt$ in $\reals^d$]
\label{lem:triple}
For three consecutive hyperballs, the expected value of of the product
of the lengths of the segments connecting random samples inside those
balls is
$$\expect[\lo\lt] \approx \clear^2 + \Big( \frac{2d-3}{d+2} \Big)\ballrad^2 $$
\end{lemma}

\subsubsection{Combining the Lemmas to compute 
$Var(\sum_{m=1}^{M} \lm)$ in $\reals^d$}

Computing the final variance now simply requires plugging in the
values from Lemmas \ref{lem:mean} to \ref{lem:triple}.  Start with
the expression computed at the beginning of this section:

\begin{multline*}
Var\big( \sum_{m=1}^{M} \lm \big) = M\expect[\lo^2]\\ + (2M-2)
\expect[\lo\lt] + (2 - 3M)\big(\expect[\lo]\big)^2 ,
\end{multline*}

and substitute the computed values:

\begin{multline*}
Var\big( \sum_{m=1}^{M} \lm \big) \approx M\Big(\clear^2 + 
\frac{2d}{d+2} \beta^2\Big)\\ + (2M-2)\Big(\clear^2 + 
\frac{2d-3}{d+2}\beta^2\Big)\\ + (2-3M)\Big(\clear + 
\frac{(d-1)\beta^2}{(d+2)\clear}\Big)^2 
\end{multline*}

\begin{multline*}
Var\big( \sum_{m=1}^{M} \lm \big) \approx M\clear^2 + \frac{2Md}{d+2}
\beta^2 + 2M\clear^2\\ + \frac{4Md-6M}{d+2}\beta^2 - 2\clear^2 - 
\frac{4d-6}{d+2}\beta^2 + 2\clear^2\\ + \frac{4d-4}{d+2}{\ballrad}^2 - 
3M\clear^2 - \frac{6Md - 6M}{d+2}{\ballrad}^2
\end{multline*}

\begin{multline*}
Var\big( \sum_{m=1}^{M} \lm \big) \approx \frac{{\ballrad}^2}{d+2} \Big( 2Md +
4Md - 6M -4d\\ + 6 +4d - 4 - 6Md + 6 \Big)
\end{multline*}

$$ Var\big( \sum_{m=1}^{M} \lm \big) = Var(\lalg) \approx \frac{2{\ballrad}^2}{d+2} $$

\begin{lemma}[Variance of $\lalg$]
$$ Var(\lalg) \approx \frac{2\ballrad^2}{d+2} $$
\end{lemma}

\begin{figure}
  \centering
  \begin{tabular}{| c | c | c |}
    \hline
    Euclidean & $\lan = 0.5$ & $\lan = 0.125$ \\
    \hline
    dimension & $\%$ error & $\%$ error \\
    \hline
    2 & 6.0245\% & 0.5739\% \\
    3 & 9.7691\% & 1.0655\% \\
    10 & 19.0989\% & 2.1429\% \\
    100 & 23.7279\% & 2.8191\% \\
    \hline
  \end{tabular}
  \vspace{-0.11in}
  \caption{Simulation comparison for $Var(\lalg)$, using 120,000 
  data points for each entry, where the error is $100\cdot
  \frac{|Var(\lalg) - Var^{MC}|}{Var^{MC}}$.}
  \label{tab:var_tight}
\end{figure}
Monte Carlo simulations are used to verify that the drawn approximation
of the variance characterizes the variance properly.  The relative
error of the variance is higher than for the mean; however, for small
values of $\lambda$, the approximation becomes tighter.  
Interestingly, up to a second-order Taylor Approximation, the variance
relies only on $\ballrad$, and not on the length of the optimal path,
\lopt.

\subsection{Finalizing the \pno\ guarantee of \prmstar}

Now that the mean and variance of $\lalg$ has been approximated,
the derivation of the bound can continue.  Recall that in Section
\ref{sec:chebyshev}, the bound was manipulated into the following
form:

\begin{multline*}
\pr\big( |\lalg - \lopt| \geq \delta\cdot\lopt \ |\ \cover  \big)\\ = 
2\pr \big( \zmu > (\delta+1)\lopt - \expect(\lalg)\ |\ \cover \big) 
\end{multline*}

\noindent Now, substituting the computed values into this expression:

$$2\pr\big( \zmu > (\delta+1)\balls\clear - \balls\big(\clear + 
\frac{(d-1)\ballrad^2}{(d+2)\clear}\big)\ |\ \cover \big) $$

$$= 2\pr\big( \zmu > \balls\clear\big(\delta - 
\frac{(d-1)\ballrad^2}{(d+2)\clear^2}\big)\ |\ \cover \big) $$

\noindent Recall that the inequality leveraged is Chebyshev's
Inequality:

$$\pr(|X-\expect[X]| \geq a)\leq \frac{Var(X)}{a^2} $$

\noindent then application of this inequality yields:

\begin{multline*}
\pr\big( |\lalg -\lopt| > \delta\cdot\lopt\ |\ \cover \big) 
\leq \ugly,\\ \text{where } \ugly =  \frac{ \frac{4\lan^2\clear^2}{d+2} }
{{\lopt}^2\big(\delta - \frac{(d-1)}{(d+2)}\lan^2 \big)^2},
\end{multline*}

\noindent where $\lan = \frac{\ballrad}{\clear}$.  Then, the 
unconditional probability can be bounded as:

$$ \pr\big( |\lalg - \lopt| \geq \delta\cdot\lopt \big) \leq 
\ugly \pr( \cover ) + (1-\pr( \cover )) $$

$$ \pr\big( |\lalg - \lopt| \geq \delta\cdot\lopt \big) \leq 
1 + \pr\big(\cover\big)(\ugly - 1) $$

\noindent This leads to the following Theorem:

\begin{theorem}[Probabilistic Near-Optimality of \prmstar]
\label{thm:pnoprm}
For finite iterations $n$, \prmstar\ is probabilistically near-optimal,
building a graph containing a path of length $\lalg$ such that
$$ \pr\big( |\lalg - \lopt| \geq \delta\cdot\lopt \big) \leq 
1 + \pr\big(\cover\big)(\ugly - 1) $$
\end{theorem}

This bound involves several variables, many of which are known.  
For instance, the path-length bound $\delta$ is required as input.  
The parameter $\balls$ is a function of the length of the optimal 
path $\lopt$ and $\clear$.  
The length of the optimal path is not known in general, so a 
pessimistic estimate is required when leveraging this guarantee.  To 
get such an estimate, if \prmstar\ has returned a solution, its length 
can be used as an estimate for $\lopt$.  This bound also depends on 
the radius of the hyperballs, $\ballrad$, and on the number of 
samples generated in $\cfree$, $n$.

\subsection{Extending \pno\ to roadmap spanners.}

The roadmaps created with asymptotically optimal planners can be
prohibitively large for practical use.  Roadmap spanners have been
proposed as practical methods for returning high-quality solutions
while reducing memory requirements \cite{Marble2013ANOJournal}.  
These methods provide the property of
asymptotic near-optimality, i.e. as the algorithm runs to infinity,
the probability that they return a path no more than $t$ times the
optimal converges to $1$.   The Sequential Roadmap Spanner (\srs) 
method performs a roadmap spanner technique over the resulting roadmap
of \prmstar, ensuring that paths generated by \srs\ are no more than
$t$ times longer than corresponding \prmstar\ paths.  This parameter
is known as the stretch of the spanner, and is taken as input to the
method.

This method was extended to work in an incremental fashion as well, 
known as the Incremental Roadmap Spanner (\irs) method.  Both \srs\ 
and \irs\ ensure as an invariant that paths returned do not violate
the stretch $t$, but consider all of the same samples and connections
that \prmstar\ would.  This leads to the following Lemma:

\begin{corollary}[Probabilistic Near-Optimality of \srs\ and \irs]
For finite iterations $n$ and input stretch $t$, \srs\ and \irs\
are probabilistically near optimal, constructing a path of length
\lspan\ such that
$$ \pr\big( |\lspan - t\cdot \lopt| \geq \delta\cdot t\cdot\lopt 
\big) \leq 1 + \pr\big(\cover\big)(\ugly - 1) $$
\end{corollary}

\section{Using \pno\ properties in practice}

The derived probabilistic near-optimality guarantee of \prmstar\ can
be leveraged in several useful ways in practice.  This section shows
how it can be used to estimate the length of the optimal path solving 
a query in the same homotopic class as the current solution during 
runtime.  It can also provide an automated stopping criterion which 
probabilistically guarantees high-quality paths as well.

\subsection{Online Prediction of $\lopt$}

The guarantee can be leveraged to make a prediction of the length
of the optimal path which answers a query during online execution of 
the algorithm within a confidence bound $\success = \pr( |\lalg - 
\lopt| < \delta\cdot\lopt )$.  A practical method for creating the 
estimate of $\lopt$ would be to consider $\balls$ for the given 
$\clear$ and of the current returned path length from the algorithm, 
$\lalg$, and set $\lopt = \frac{\lalg}{(\delta + 1)}$.  Recall that:

$$ \pr\big( |\lalg - \lopt| \geq \delta\cdot\lopt \big) \leq 
1 + \pr\big(\cover\big)(\ugly - 1) $$

\noindent Furthermore, it was shown that:

$$ \pr\big( |\lalg - \lopt| \geq \delta\cdot\lopt \big) =
2\pr\big( \lalg - \lopt \geq \delta\cdot\lopt \big), $$

$$ \pr\big( |\lalg - \lopt| \geq \delta\cdot\lopt \big) \geq
\pr\big( \lalg - \lopt \geq \delta\cdot\lopt \big), $$

\noindent and it must also be that:

\begin{multline*}
\pr\big( \lalg - \lopt \geq \delta\cdot\lopt \big) =
\pr\big( \lopt \leq \frac{\lalg}{\delta + 1} \big) \leq \\
\pr\big( |\lalg - \lopt| \geq \delta\cdot\lopt \big) \leq
1 + \pr\big(\cover\big)(\ugly - 1)
\end{multline*}

\noindent Consider, however, that this result is only valid given
that $\piopt$ exists for the current value of $\clear$.  Therefore,
it is critical that the algorithm executes at least until $\clear
\leq \enaught$, i.e. when $n \geq \nnaught$.  Then all that remains 
is to solve the bound in terms of $\delta$.  It is known that

$$ \pr\big( \lopt \leq \frac{\lalg}{\delta + 1} \big) \leq 
1 + \pr\big(\cover\big)(\ugly - 1) \geq 1 - \success  $$

\noindent Then, the goal is to solve for $\delta$.  Performing some 
algebraic manipulation:

$$ \ugly \geq 1 - \frac{\success}{\pcover} $$

\noindent Then, substituting the value for $\ugly$ yields,

$$ \frac{ \frac{4\lan^2\clear^2}{d+2} }
{{\lopt}^2\big(\boundn - \frac{d-1}{d+2} \lan^2 \big)^2} 
\geq 1 - \frac{\success}{\pcover} $$

$$ \frac{4\lan^2\clear^2}{(1- \frac{\success}{\pcover})(d+2){\lopt}^2}
\geq \big(\boundn - \frac{(d-1)}{(d+2)}\lan^2 \big)^2 $$

$$ \sqrt{\frac{4\lan^2\clear^2}{(1- \frac{\success}{\pcover})(d+2)
{\lopt}^2}} + \frac{(d-1)}{(d+2)}\lan^2 \geq \boundn $$

$$\boundn \leq \frac{2\lan\clear}{\lopt}\sqrt{\frac{1}{(d+2)(1 - 
\frac{\success}{\pcover})}} + \frac{(d-1)}{(d+2)}\lan^2 $$

\begin{lemma}[Multiplicative bound $\boundn$]
After $n > \nnaught$ iterations of \prmstar, with probability 
\success, if \piopt\ exists, then \prmstar contains a path 
$\boundn$-bounded by \lopt\ where:
\begin{equation}
\boundn \leq \frac{2\lan\clear}{\lopt}\sqrt{\frac{1}{(d+2)(1 - 
\frac{\success}{\pcover})}} + \frac{(d-1)}{(d+2)}\lan^2 
\label{eq:boundn}
\end{equation}
\end{lemma}

\subsection{Deriving a probabilistic stopping criterion}

One helpful use of this guarantee is to set a desired confidence 
probability \pdes\ of returning a path within a desired quality bound 
$\ddes$.  For input $\enaught$, there exists some 
$\enaught$-robust optimal path, \pinaught\ of length \lnaupt.
Then, using Equations \ref{eq:pcover} and \ref{eq:boundn}, it is 
possible to compute a required iteration $\nnaught$, such that a path 
\pinaught\ covering \pinaupt\ has been computed which has length
\lnaught\ bounded by $\ddes$ with probability \pdes.
The limit will be derived using Equations
\ref{eq:boundn} and \ref{eq:orig_pcover}.  First, let $\lan = 
\frac{1}{2}$ and then manipulate Equation \ref{eq:boundn} to solve 
for $\pcover$:

$$ \boundn \leq \frac{1}{\balls} \sqrt{\frac{1}{(d+2)\big( 1 - \frac
{\success}{\pcover} \big)}} + \frac{1}{4}\cd  $$

$$ \balls\big(\boundn - \frac{1}{4}\cd\big) \leq \sqrt{\frac{1}
{(d+2)\big( 1 - \frac{\success}{\pcover} \big)}} $$

$$ \balls^2\big(\boundn - \frac{1}{4}\cd\big)^2 \leq \frac{1}
{(d+2)\big( 1 - \frac{\success}{\pcover} \big)} $$

$$(d+2)\big( 1 - \frac{\success}{\pcover} \big) \geq \frac{1}
{\balls^2\big(\boundn - \frac{1}{4}\cd\big)^2} $$

\noindent Then, finally solving for $\pcover$, the right hand side 
will be denoted as $\ougly$:

$$ \pcover \geq \frac{1}{\frac{1}{\pdes}\cdot\Big( 1 - \frac{1}
{\ballnaught^2 \cdot (d+2)\big(\ddes - \frac{(d-1)}{4(d+2)}\big)^2} 
\Big)} = \ougly $$

Then, substituting the form of Equation \ref{eq:orig_pcover} using
$\beta_0$, $\ballnaught$, and $\nnaught$ yields:

$$ \bigg(1 - \Big(1 - \frac{|\ball_{\beta_0}|}{|\cfree|} \Big)^{\nnaught}
\bigg)^{\ballnaught} \geq \ougly $$

\noindent Solving for $\nnaught$:

$$ -\Big( 1 - \frac{|\ball_{\beta_0}|}{|\cfree|} \Big)^{\nnaught} \geq 
\sqrt[\ballnaught]{\ougly} - 1 $$

$$ \Big( 1 - \frac{|\ball_{\beta_0}|}{|\cfree|} \Big)^{\nnaught} \leq 
1 - \sqrt[\ballnaught]{\ougly}  $$

$$ \nnaught \leq \Bigg\lceil \frac{\log{(1 - \sqrt[\ballnaught]{\ougly})}}
{\log{(1 - \frac{|\ball_{\beta_0}|}{|\cfree|})}} \Bigg\rceil$$

\noindent Here, $\nnaught$ represents a maximum number of samples 
\prmstar\ must be run in order to guarantee \pno\ properties.

\begin{lemma}[\pno\ iteration limit for \prmstar]
For given \ddes\ and \pdes, the graph of \prmstar\ probabilistically 
contains a path \pinaught\ of length \lnaught\ with $\pr\big( 
|\lnaught - \lnaupt| \geq \ddes \cdot \lnaupt \big) \leq 1 - \pdes$
after $\nnaught$ iterations, where
\begin{multline}
\nnaught \leq \Bigg\lceil \frac{\log{(1 - \sqrt[\ballnaught]{\ougly})}}
{\log{(1 - \frac{|\ball_{\beta_0}|}{|\cfree|})}} \Bigg\rceil, 
\text{where,} \\ \ougly = \frac{1}{\frac{1}{\pdes}\cdot\Big(1 - \frac{1}
{\ballnaught^2 \cdot (d+2)\big(\ddes - \frac{(d-1)}{4(d+2)}\big)^2} 
\Big)}
\label{eq:nnaught}
\end{multline}
\end{lemma}

\subsection{Considering Non-Euclidean spaces}

The derivation of the \pno\ guarantee assumed that the distance
between points in the space and the length of paths is expressed in
terms of the $L_2$ norm in Euclidean d-space.  \pno\ guarantees can 
be drawn for most relatively well behaved  metric spaces, though this 
requires deriving $\expect[\lalg]$, $\expect[\lo^2]$, and 
$\expect[\lo\lt]$ for the particular metric examined.  The drawn 
guarantee can still be used for any space where the $L_2$ norm is 
applicable, at least in a local sense, i.e. the space is locally 
homeomorphic to a d-dimensional Euclidean space.

\section{Indications from Simulation}

\begin{figure}
\centering
\includegraphics[width=2.62in]{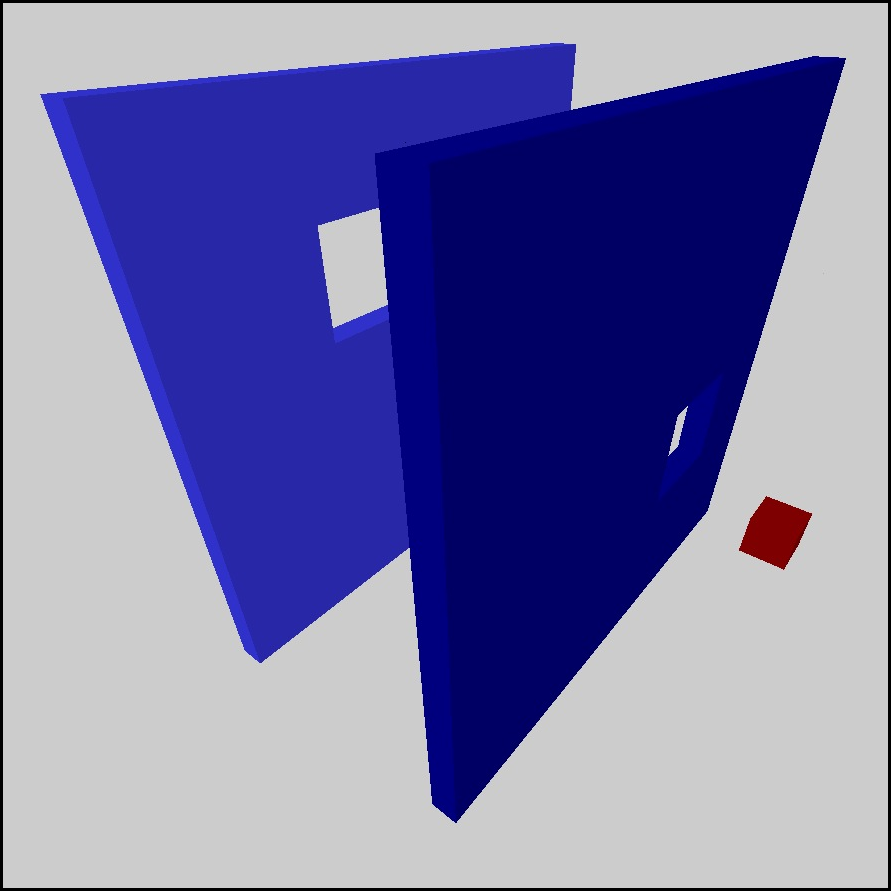}
\caption{The setup for validating the stopping criterion, for a 3D 
rigid body without rotations.}
\label{fig:environs}
\end{figure}
To test the validity of the analysis, experiments were performed using
the described \prmstar\ in the environment illustrated
in Figure \ref{fig:environs}.  Experiments were run using the PRACSYS
Library \cite{Kimmel2012PRACSYS:-An-Ext}.  The automated stopping 
criterion was tested to ensure it stops \prmstar\ appropriately.  
The automated stopping criterion should stop the algorithm at an 
iteration when the \pno\ condition is satisfied, and not let the
algorithm run excessively longer than required.

\begin{figure}
\centering
\includegraphics[width=2.86in]{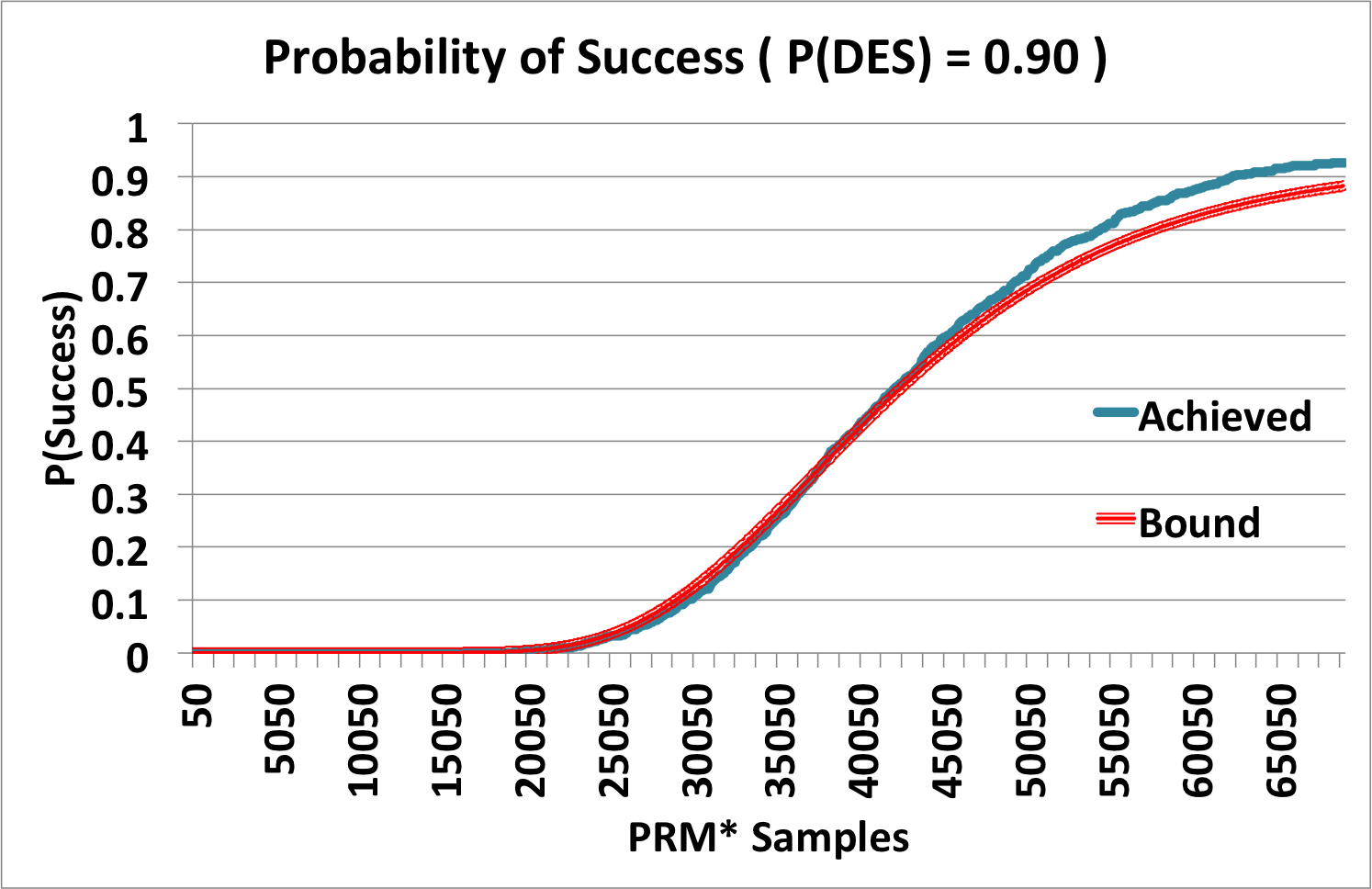}
\includegraphics[width=2.86in]{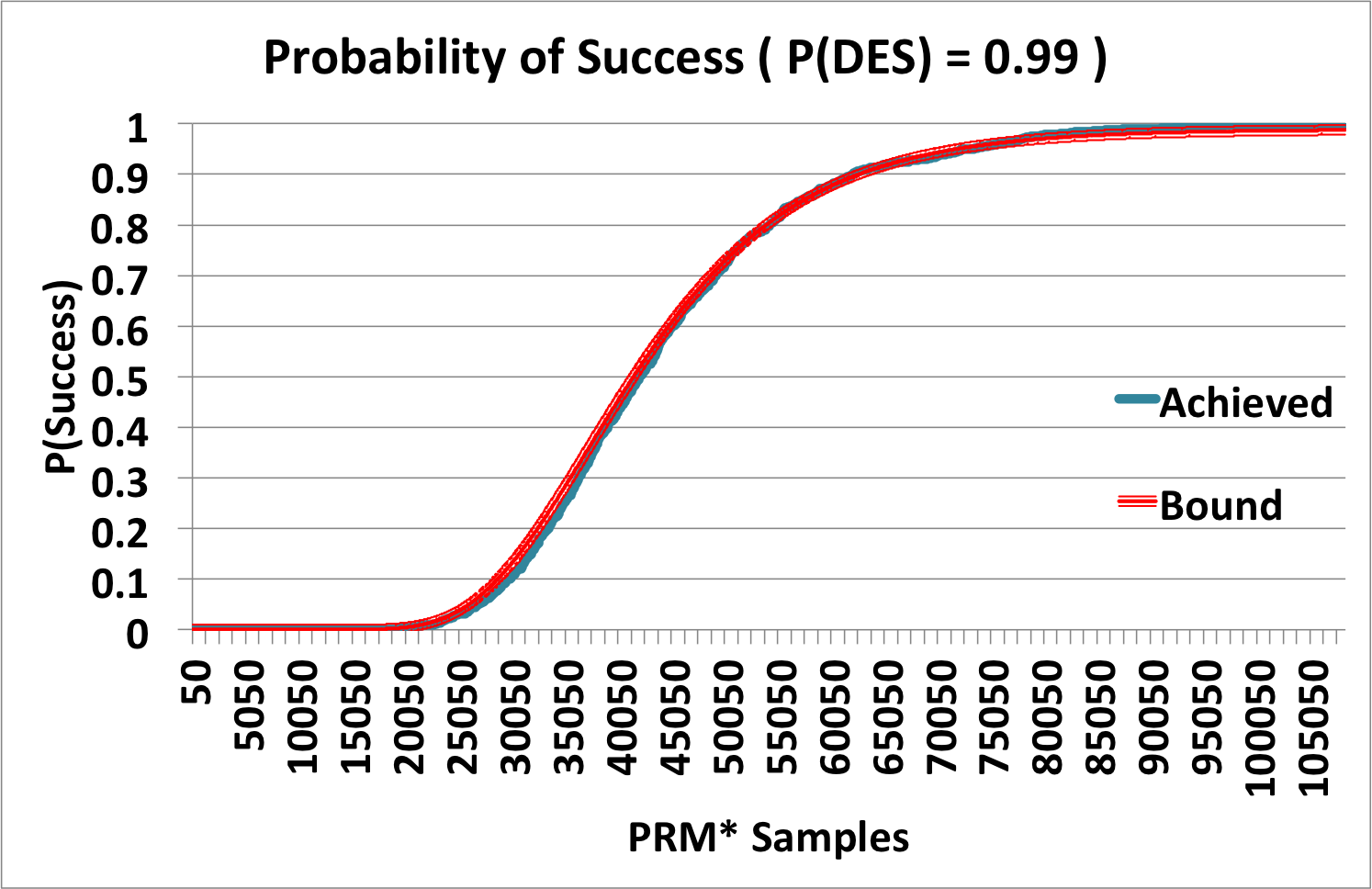}
\caption{Probability of successfully returning a path within
the bound \ddes\ for \prmstar\ over time.}
\label{fig:graphs}
\end{figure}

The results of running the stopping criterion are summarized in 
Figure \ref{tab:validation}.  For the desired path bound and
probability of success, the iteration limit \nnaught\ was 
computed.  Then, out of 1000 experimental trials, the actual
probability of successfully generating a path through the set
of hyperballs over \pinaught\ is computed.  The stopping criterion
properly selects \nnaught\ so that the \success\ is greater than
the input threshold \pdes.  The probability of success for the 
algorithm over time is given in Figure \ref{fig:graphs} for the 
two settings of \pdes.  

\begin{figure}
  \centering
  \begin{tabular}{| c | c | c | c | c |}
    \hline
    \enaught & \nnaught & \ddes & \pdes & \success \\
    \hline
    0.5 & 69429 & 0.16 & 0.9 & 0.93 \\
    0.5 & 108328 & 0.25 & 0.99 & 0.998 \\
    \hline
  \end{tabular}
  \caption{Results for the 3D Rigid Body scenario.  The automated
  stopping criterion terminates execution to appropriately ensure
  $\success \geq \pdes$. }
\label{tab:validation}
\end{figure}

\section{Discussion}

This work formally shows \pno\ properties for an asymptotically
optimal sampling-based planner, overcoming limitations of prior work.
The new framework shows \pno\ properties using an asymptotically
sparser planning structure, and removes dependence on Monte Carlo
simulations.  The analysis shows tight bounds for path quality, and
experimental results show that these properties practically guarantee
high-quality solutions in finite time.


There are many avenues for future investigation.  An important step is
to ensure that \pno\ properties can be extended to the tree-based
planner \rrtstar.  Furthermore, these methods still require a large
amount of samples, so it is pertinent to determine how
\pno\ properties can be extended to roadmap spanner techniques which
remove nodes from the planning structure
\cite{Dobson2014Sparse-Roadmap}.  The drawn bounds reason over a
single path which exists in the planning structure, however, less
conservative bounds can be drawn if many paths can be considered
simultaneously. It is also interesting to see if \pno\ properties can
be leveraged to better inform task planners of problem difficulty
along different exploration directions.  Furthermore, the approach
will become more broadly applicable if the bound can be generalized
for non-$L_2$ norms.  The drawn bounds can also be improved, for
instance, by computing an analytical solution instead of the
approximate bounds computed here, by considering the effects of having
multiple samples in each hyperball, or by finding tighter bounds where
Chebyshev's Inequality was employed. An interesting prospect is to
investigate low-dispersion sampling approaches, which can effectively
ensure that the algorithm always generates samples within the
hyperballs.

{\small
\bibliography{references}
\bibliographystyle{abbrvnat}
}


\end{document}